\documentclass[11pt, a4paper]{deepmind}

\usepackage[authoryear, sort&compress, round]{natbib}
\usepackage{microtype}
\usepackage{graphicx}
\usepackage{subfigure}
\usepackage{booktabs} 
\usepackage{listings}
\usepackage{algorithm}
\usepackage{algorithmic}

\usepackage{hyperref}

\newcommand{\cS}{\mathcal{S}}
\newcommand{\cA}{\mathcal{A}}
\newcommand{\cH}{\mathcal{H}}

\newcommand{\cR}{\mathcal{R}}
\newcommand{\cO}{\mathcal{O}}
\newcommand{\EE}{\mathbb{E}}

\usepackage{amsmath,amsfonts,bm}

\def\eqref#1{equation~\ref{#1}}

\def\1{\bm{1}}

\DeclareMathAlphabet{\mathsfit}{\encodingdefault}{\sfdefault}{m}{sl}
\SetMathAlphabet{\mathsfit}{bold}{\encodingdefault}{\sfdefault}{bx}{n}

\newcommand{\E}{\mathbb{E}}

\newcommand{\R}{\mathbb{R}}

\DeclareMathOperator*{\argmax}{arg\,max}

\newcount\Comments
\newcommand{\kibitz}[2]{\ifnum\Comments=1{\textcolor{#1}{\textsf{\footnotesize #2}}}\fi}

\newcommand{\env}{\texttt{Env}}
\newcommand{\agent}{\texttt{Agent}}
\newcommand{\pinit}{p_{\text{init}}}
\newcommand{\bsuite}{\texttt{bsuite}}

\usepackage{amsmath}
\usepackage{amssymb}
\usepackage{mathtools}
\usepackage{amsthm}
\usepackage{multirow}
\usepackage[capitalize,noabbrev]{cleveref}

\theoremstyle{plain}

\theoremstyle{definition}

\theoremstyle{remark}

\newcommand{\localplanning}{\textsc{UncertaintyFirstLocalPlanning}}

\def\arxiv{1}

\title{Sample Efficient Deep Reinforcement Learning via Local Planning}

\correspondingauthor{ \{dongyin, sthiagarajan, nevena\}@google.com }

\reportnumber{} 

\renewcommand{\today}{2023-06-20}

\author[*,1]{Dong Yin}
\author[*,1]{Sridhar Thiagarajan}
\author[*,1]{Nevena Lazic}
\author[2]{Nived Rajaraman}
\author[1]{Botao Hao}
\author[1]{Csaba Szepesvari}

\affil[*]{Equal contributions}
\affil[1]{Google DeepMind}
\affil[2]{UC Berkeley}

\begin{abstract}
The focus of this work is sample-efficient deep reinforcement learning (RL) with a simulator. One useful property of simulators is that it is typically easy to reset the environment to a previously observed state. We propose an algorithmic framework, named \emph{uncertainty-first local planning} (UFLP), that takes advantage of this property. Concretely, in each data collection iteration, with some probability, our meta-algorithm resets the environment to an observed state which has high uncertainty, instead of sampling according to the initial-state distribution. The agent-environment interaction then proceeds as in the standard online RL setting. We demonstrate that this simple procedure can dramatically improve the sample cost of several baseline RL algorithms on difficult exploration tasks. Notably, with our framework, we can achieve super-human performance on the notoriously hard Atari game, Montezuma's Revenge, with a simple (distributional) double DQN. Our work can be seen as an efficient approximate implementation of an existing algorithm with theoretical guarantees, which offers an interpretation of the positive empirical results.
\end{abstract}

\begin{document}

\maketitle

\section{Introduction}\label{sec:intro}

Simulators are ubiquitous in modern reinforcement learning (RL). They correspond to either to the environment itself (as in chess, go, and video games~\citep{bellemare2013arcade}) or to a simplified model of the true environment (such as robotic arm manipulation \citep{qassem2010modeling,akkaya2019solving}, car driving \citep{bojarski2016end,aradi2020survey}, or plasma shape control in fusion \citep{degrave2022magnetic}).
Simulators have been widely used in RL research. Many standard benchmarks in RL involve simulators, for example, Atari games~\citep{bellemare2013arcade}, Mujoco simulation engine~\citep{todorov2012mujoco}, OpenAI Gym~\citep{brockman2016openai}, DeepMind control suite~\citep{tassa2018deepmind}, and DeepMind Lab~\citep{beattie2016deepmind}.
Somewhat surprisingly, the majority of RL algorithms use the agent-environment interaction protocols that mimic learning in the real world during training, and do not explicitly take advantage of favorable simulator properties. In particular, the standard interaction protocol, called \emph{online access}, assumes that the agent can only follow the dynamics of the environment during learning. In this work, we consider the \emph{local access}  protocol \citep{yin2021efficient}, where the agent is allowed to revisit any previously observed state in addition to following the dynamics. This protocol can easily be implemented with simulators for many commonly used RL environments (see, e.g., Appendix~\ref{sec:checkpointing}).

Local access has received less attention from the RL community compared online access. On the theory side, several recent works show that local access makes sample-efficient learning possible in settings where it has not been shown in the online access setting \citep{li2021sample, yin2021efficient, hao2021confident}. On the empirical side, the \emph{vine} method in TRPO~\citep{schulman2015trust} uses local access to obtain better estimates of the value function, and the Go-Explore algorithm of \citet{ecoffet2019go} relies on local access to achieve state-of-the-art performance on several hard-exploration Atari games. Intuitively, the main advantage of local access is that we can directly reset the simulator to the states that can provide more information to the agent, and thus improve the exploration of the state space.

\paragraph{Contributions} Our contribution is three-fold:
\ifx\arxiv\undefined
\vspace{-0.1in}
\fi
\begin{itemize}
    \item We propose a general algorithmic framework for RL with a simulator under the local access protocol. Our framework, named \emph{uncertainty-first local planning} (UFLP), revisits states from the agent's history based on the uncertainty about their value.

    \item We instantiate this framework with several base RL agents (deep Q-networks, policy iteration) and uncertainty estimates (ensemble, feature covariance, approximate counts, random network distillation).

    \item We demonstrate that UFLP can significantly improve the sample cost compared to online access on difficult exploration tasks, including the Deep Sea and Cartpole Swingup benchmarks in \texttt{bsuite} \citep{osband2019behaviour} and hard-exploration Atari games: Montezuma's Revenge and PrivateEye. In particular, for the Deep Sea environment, by leveraging UFLP, many RL agents can easily solve the task, whereas in the online access setting they can only obtain zero reward. For Montezuma's Revenge, applying UFLP on top of a simple double DQN \citep{van2016deep} results in a super-human score.
\ifx\arxiv\undefined
\vspace{-0.1in}
\fi
\end{itemize}
Our work also opens up a new research avenue for improving sample efficiency when learning with simulators.

\section{Related Work}\label{sec:related}

\paragraph*{Local access protocol} Simulators are routinely used in RL algorithms that leverage Monte Carlo tree search~\citep{coulom2006efficient,kocsis2006bandit}, including prominent examples such as AlphaGo~\citep{silver2016mastering} and AlphaZero~\citep{silver2018general}. One major difference between the tree-search setting and the setting that we consider in this paper is that the tree-search algorithms usually make the assumption that the agent has local access to the simulator during \emph{evaluation}, whereas we only consider local access during training. This means that once the training is finished, the agent is evaluated against the environment without access to the simulator. Therefore, our framework is more suitable for applications that require fast inference when the agent is deployed. Moreover, most tree-search approaches do not select states to revisit strategically; instead, they expand the search tree based on visitation counts, rather than more general uncertainty metrics.

One notable example of RL with local access protocol is the Go-Explore algorithm \citep{ecoffet2019go,ecoffet2021first}. This algorithm operates under the local access protocol and revisits states deemed to be ``promising'' in history.
It then uses backward learning-from-demonstrations \citep{salimans2018learning} to learn a robust policy. While Go-Explore achieves or surpasses the state of the art on $11$ Atari games, the algorithm design, especially the state revisiting rule, uses heuristics tailored to these games, and it is unclear how to combine this method with other uncertainty metrics or more general RL agents. By contrast, in this paper, we propose a \emph{general algorithm framework} that can be combined with most existing RL agents in order to improve their performance.

Besides Go-Explore, a few other prior works have studied the use of local access. In an early literature on real-time dynamic programming~\citep{barto1995learning,mcmahan2005bounded,smith2006focused,sanner2009bayesian}, it has been shown that prioritizing revisiting states with higher uncertainty is helpful for achieving faster convergence. However, these works mainly focus on tabular MDPs and only consider the value iteration algorithm. On the contrary, our general framework is suitable for function approximation and can be combined with other types of agents beyond value iteration.
In the vine method in the TRPO algorithm~\citep{schulman2015trust}, local access is used to obtain more accurate estimates of the value function. This differs from our work since we focus on improving exploration of deep RL agents. Restart distributions have also been explored in \citet{tavakoli2018exploring}, who consider revisiting states uniformly at random, according to TD error, and according to episode returns, in combination with the PPO algorithm \citep{schulman2017proximal}.
Revisiting based on high episode returns improves the performance of PPO on a sparse-reward task deemed hard-exploration. However, it is unclear that the same approach would be successful in environments such as Deep Sea~\citep{osband2019behaviour}, where discovering an episode with a positive return requires non-trivial exploration. A very recent work by~\citet{lan2023can} focuses on generalization of RL agents to out-of-distribution trajectories using local access to a simulator. Their algorithm can be considered as a special case in our framework.

On the theoretical side, several recent works have proposed sample- and computationally-efficient algorithms under the local access protocol and with linearly-realizable action-value functions \citep{li2021sample,yin2021efficient,hao2021confident,weisz2022confident}. The works of \citet{yin2021efficient}, \citet{hao2021confident}, and \citet{weisz2022confident} maintain a \emph{core set} of previously-visited states that cover different parts of the feature space.
In the online access setting, a similar idea is used in the policy cover policy gradient (PCPG) algorithm of \citet{agarwal2020pc} and the follow-up work of \citet{zanette2021cautiously}.
Our work is motivated by the approaches of \citet{yin2021efficient} and \citet{hao2021confident}, but adapted to the practical setting of value learning with neural network function approximation, where the core set is more difficult to define rigorously. 

\paragraph*{Uncertainty estimation and exploration in RL}~~Successful methods for learning in MDPs typically rely on estimates of uncertainty about the value of state-action pairs in order to encourage the agent to explore the environment.  One type of exploration strategies rely on uncertainty-based \emph{intrinsic rewards} or \emph{bonuses}.
Uncertainty metrics based on feature covariance have been used in theoretical RL works with linear function approximation~\citep{jin2020provably}. Empirically, popular approaches include approximate count~\citep{tang2017exploration,bellemare2016unifying}, random network distillation (RND)~\citep{burda2018exploration}, and curiosity-driven exploration~\citep{pathak2017curiosity}.
Recent successful approaches \citep{badia2020never, badia2020agent57} have constructed bonuses based on nearest-neighbors in the current episode, as well as RND to capture longer-term uncertainty. Another set of approaches rely on randomized value functions \citep{osband2016deep,osband2018randomized}. We discuss these methods in more detail in Section~\ref{sec:base_agents}.

\section{Problem Setting}
We use $\Delta_{\cS}$ to denote the set of probability distributions defined on any countable set $\cS$ and write $[N] := \{1,2, \ldots, N\}$ for any positive integer $N$.

An infinite-horizon discounted Markov decision process (MDP) can be characterized by a tuple $(\cS, \cA, R, P, \mu_0, \gamma)$, where $\cS$ is the state space, $\cA$ is the action space, $R : \cS \times \cA \rightarrow [0, 1]$ is the reward function, $P : \cS \times \cA \rightarrow \Delta_\cS$ is the probability transition kernel, $\mu_0$ is the initial state distribution, and $\gamma \in (0, 1)$ is the discount factor. Both $P$ and $R$ are unknown. In this paper, we only consider finite action space $|\cA| < \infty$.

At each state $s$, if the agent picks an action $a\in\cA$, the environment evolves to a random next state $s'$ according to the distribution $P(s'|s,a)$ and generates a stochastic reward $r\in[0, 1]$ with $\mathbb E[r|s,a] = R(s,a)$.

A stationary policy $\pi : \cS \rightarrow \Delta_\cA$ is a mapping from a state to a distribution over actions. For a policy $\pi$, its value function $V_\pi(s)$ is the expectation of cumulative rewards received under policy $\pi$ when starting from a state $s$, i.e.,
$V_\pi(s) = \E^{\pi} \left[ \sum_{t=0}^\infty \gamma^t R(s_t, a_t) \Big| s_0=s \right]$,
where $a_t \sim \pi(\cdot | s_t), s_{t+1}\sim P(\cdot | s_t, a_t)$ and $\mathbb E^{\pi}$ denotes the expectation over the sample path and stochastic reward generated under policy $\pi$.
The action value function $Q_\pi(s, a)$ is defined as
$
Q_\pi(s, a) = \E^{\pi} \left[ \sum_{t=0}^\infty \gamma^t R(s_t, a_t) \Big| s_0=s, a_0=a \right]$.

\subsection{Simulator Interaction Protocol}
We distinguish between three protocols for interacting with the MDP simulator (or environment) commonly used in the literature.
\begin{itemize}
\ifx\arxiv\undefined
\vspace{-0.1in}
\fi
\item \emph{Online access}. The initial state $s_0$ is sampled from the initial state distribution $\mu_0$. The agent can only reset the environment to a (possibly random) initial state, or move to the next state given an action by following the MDP dynamics. 

\item \emph{Local access}. The agent can reset the environment to a random initial state, or to a state that has previously been observed. 

\item \emph{Random access}. The agent can query the simulator with \emph{any} state-action pair of its choice to obtain a reward and a sample of the next state. This is often referred as the \emph{access to a generative model} in literature \citep{kakade2003sample, sidford2018near, yang2019sample}.
\ifx\arxiv\undefined
\vspace{-0.1in}
\fi
\end{itemize}

Most RL algorithms use the online access protocol, which also mimics learning in the real world. Random access is primarily considered in theoretical works, as it enables sample-efficient learning in settings where this is impossible under online access~\citep{du2019good}. Unfortunately, this interaction protocol is often difficult or impossible to support in large-scale MDPs where the agent may not even know which states exist or are plausible. For example, for random access, the agent would need to know which positions and velocities of a robotic arm are valid according to physics, or which images correspond to a valid frame of a video game.
The local access protocol does not suffer from this issue, as the agent is only allowed to revisit previously observed states which are known to be plausible. Local access is also easy to implement in most simulators, for example by checkpointing the simulator state.

The intuition on why using local access can improve sample efficiency of policy optimization is that we can directly reset the simulator to the states that can provide more information to the agent. In other words, we can directly start data collection from the states with high uncertainty. In the online access mode, exploration methods such as additive bonus and Thompson sampling~\citep{thompson1933likelihood,osband2016deep} have been designed to achieve the similar goal; however, in this setting, the agent still has to start from the initial state, reach an uncertain state, and then collect data there. Therefore, local access saves the sample cost by leveraging the resetting ability of simulators.
\section{Algorithm Framework}

In this section, we present an algorithm framework for policy optimization with local access to a simulator. There are four main components in our framework:
\begin{itemize}
\item A \emph{simulator} (or environment) that we can reset to any state that has been observed during the learning process, denoted by \env~in the following. We denote the operation of resetting the environment to a given observed state $s$ by $\env.\textsc{Reset}(s)$. We also denote the operation of stepping the environment (taking an action and moving to the next state) by $\env.\textsc{Step}(a)$.
\item A \emph{base agent} (\agent) that can take actions given the observation of a state $s$ ($\agent.\textsc{Act}(s)$) and update itself ($\agent.\textsc{Update}()$) given collected data. In fact, any agent for online-access RL can be used as a base agent in our framework.
\item A \emph{function} $u:\cS\times\cA\mapsto \R$ that measures the \emph{uncertainty} of the agent about the value of state-action pairs. In some cases, we only define the uncertainty of the states, i.e., $u:\cS\mapsto \R$. This function is are typically updated during learning.
\item A \emph{history buffer} $\cH$. Each element in $\cH$ contains all the necessary information to reset the environment to a particular state $s$. Note that the history buffer differs from the replay buffer, which is usually used to maintain state-action transition tuples and update the agent. In the following, we omit the role of the replay buffer and mainly focus on the use of the history buffer.
\end{itemize}

Similarly to online access, our framework includes a data collection process, where the agent interacts with the environment and collects data, and a learning process where the agent is updated. The major difference in our framework is that we need to specify a starting state-action pair in the data collection process; more specifically, we reset the simulator to a given state, take a given action, move to the next state and follow the agent's action selection afterwards. This process, denoted by $\textsc{DataCollection}(\env, \agent, s_0, a_0, \cH)$ is described in Algorithm~\ref{alg:data_collection}.

\begin{algorithm}
\caption{$\textsc{DataCollection}(\env, \agent, s_0, a_0, \cH)$}
{\bf Input}: environment \env, base agent \agent, starting state-action $s_0,a_0$, history buffer $\cH$.

\begin{algorithmic}
\STATE $\cH\leftarrow \cH\cup\{s_0\}$
\STATE $\env.\textsc{Reset}(s_0)$
\STATE $s\leftarrow \env.\textsc{Step}(a_0)$
\WHILE{end of episode not reached}
\STATE $\cH\leftarrow \cH\cup\{s\}$
\STATE $a\leftarrow \agent.\textsc{Act}(s)$
\STATE $s\leftarrow \env.\textsc{Step}(a)$
\ENDWHILE
\end{algorithmic}
\label{alg:data_collection}
\end{algorithm}

With these components, we are ready to present our algorithm framework, \emph{uncertainty-first local planning} (UFLP). Here, we use the term \emph{planning} to distinguish our learning setting from the online access mode where data must be collected episode-by-episode during training.
In UFLP, in each data collection iteration, with probability $\pinit\in[0, 1]$, we sample an initial state $s_0$ according to the initial state distribution $\mu_0$ and start data collection from $s_0$. Otherwise, we sample a batch of $B$ elements from the history buffer $\cH$, denoted by $\cH_B$, pair these states with all possible actions, and choose the highest-uncertainty state-action pair as the starting point, i.e., we choose the starting point according to
\begin{align}\label{eq:most_uncertain}
\ifx\arxiv\undefined
\vspace{-0.1in}
\fi
    s_0, a_0 \leftarrow \argmax_{s \in \cH_B, a \in \mathcal{A}} u(s, a).
\ifx\arxiv\undefined    
\vspace{-0.1in}
\fi
\end{align}
As we can see, if $\pinit=1$, the algorithm reduces to the online access mode. We present details of our framework in Algorithm~\ref{alg:local_access}, where we use $\rm{Unif}[0, 1]$ to denote a random number that is sampled uniformly at random from $[0, 1]$.

\begin{algorithm}
\caption{\localplanning}
{\bf Inputs}: environment \env, base agent \agent, probability of starting from initial state $\pinit \in [0, 1]$, history buffer batch size $B$, uncertainty metric $u$.

\begin{algorithmic}
\STATE $\cH\leftarrow \emptyset$
\WHILE{termination criteria not met}
\IF {${\rm Unif}[0, 1] \leq \pinit$ or $\cH = \emptyset$}
\STATE Get $s_0\sim \mu_0$, $a_0 \leftarrow\agent.\textsc{Act}(s_0)$ 
\ELSE
\STATE Sample $B$ elements from $\cH$, denoted by $\cH_B$.
\STATE
$s_0, a_0 \leftarrow \argmax_{s \in \cH_B, a \in \mathcal{A}} u(s, a)$
\ENDIF
\STATE $\textsc{DataCollection}(\env, \agent, s_0, a_0, \cH)$
\STATE \agent.\textsc{Update}()
\ENDWHILE
\end{algorithmic}
\label{alg:local_access}
\end{algorithm}

One intuition behind the criterion that chooses an uncertain state as a starting point is that it expands the subset of the state space that we can use to start the data collection process, which in turn helps control extrapolation errors in value function estimation. Revisiting uncertain states can also improve sample efficiency in environments where states that are important for decision-making are difficult to reach. 

We also note that in practice, storing all the states that the agent has visited during training may require too much memory. Therefore, we implement the history buffer $\cH$ using a FIFO queue. Another note is that if we only have an uncertainty metric for states rather than state-action pairs, we can choose the most uncertain state in $\cH_B$ and pair it with a random action, i.e.,
\begin{align}\label{eq:most_uncertain_2}
\ifx\arxiv\undefined
\vspace{-0.1in}
\fi
s_0\leftarrow \argmax_{s \in \cH_B} u(s),~ a_0\sim\rm{Unif}(\cA).
\ifx\arxiv\undefined
\vspace{-0.1in}
\fi
\end{align}
Our experiments in Section~\ref{sec:bsuite} for \bsuite~environments~\citep{osband2019behaviour} use Eq.~\ref{eq:most_uncertain} and those in Section~\ref{sec:atari} for Atari games~\citep{bellemare2013arcade} use Eq.~\ref{eq:most_uncertain_2}.\footnote{We also experimented with Eq.~\ref{eq:most_uncertain} for Atari games. However, Eq.~\ref{eq:most_uncertain_2} led to slightly better results, and thus we report the Atari results with Eq..~\ref{eq:most_uncertain_2}.}

Next, we describe several instantiations of base agents and uncertainty metrics that can be used with the local access protocol.

\ifx\arxiv\undefined
\vspace{-0.1in}
\fi

\section{Base Agents and Uncertainty Metrics}

\subsection{Base Agents}\label{sec:base_agents}

For base agents, we consider the following commonly used ones: double deep Q network (DDQN)~\citep{van2016deep}, bootstrapped DDQN (BootDDQN)~\citep{osband2016deep,osband2018randomized}, distributional DDQN~\citep{bellemare2017distributional},  and approximate policy iteration (PI)~\citep{bertsekas2011approximate}. 

\ifx\arxiv\undefined
\textbf{DDQN}
\else
\paragraph{DDQN}
\fi
Double DQN is an improvement of the original DQN agent by~\citet{mnih2015human}. In DDQN, the agent is updated by minimizing the following loss over the transition tuples of the form $(s_t, a_t, r_t, s_{t+1})$ sampled from the replay buffer $\cR$:
\ifx\arxiv\undefined
\begin{equation}\label{eq:ddqn}
\begin{aligned}
   &L_{\text{DDQN}}(\theta) = \EE_{\cR} \big(
    Q(s_t, a_t; \theta) \\
    &- r_t - \gamma Q(s_{t+1}, \arg\max_{a\in\cA} Q(s_{t+1}, a; \theta); \theta') \big)^2,
\end{aligned}
\end{equation}
\else
\begin{equation}\label{eq:ddqn}
   L_{\text{DDQN}}(\theta) = \EE_{\cR} \big(
    Q(s_t, a_t; \theta) - r_t - \gamma Q(s_{t+1}, \arg\max_{a\in\cA} Q(s_{t+1}, a; \theta); \theta') \big)^2,
\end{equation}
\fi
where $\theta$ denotes the parameters of the Q-network, and $\theta'$ denotes the parameters of the target network that is periodically updated. During acting, one can use the standard $\epsilon$-greedy strategy, where with probability $\epsilon$, we take a random action, and otherwise we act greedily w.r.t. $Q(s, a; \theta)$.

To improve exploration, one can use an additive bonus, a.k.a. optimism. There are two common approaches. First, adding an \emph{acting-time} bonus  means that we fit the Q-network using Eq.~\ref{eq:ddqn} and select actions according to
\begin{align}\label{eq:acting_time_bonus}
\textsc{Act}(s)=\arg\max_{a\in\cA}Q(s, a; \theta) + cu(s, a),
\end{align}
where $u(s, a)$ is the uncertainty metric and $c>0$ is a scaling factor. A similar approach has been discussed in~\citet{chen2017ucb}. The second approach is to add an \emph{intrinsic reward} to the reward $r_t$ provided by the environment, i.e., replace $r_t$ in Eq.~\ref{eq:ddqn} with
\ifx\arxiv\undefined
$r'_t = r_t + cu(s_t, a_t)$
\else
\begin{align}\label{eq:intrinsic}
    r'_t = r_t + cu(s_t, a_t)
\end{align}
\fi
and train the Q-network with $r'_t$. This approach has been widely used in the literature~\citep{tang2017exploration,bellemare2016unifying,badia2020never}. In the following, we call the DDQN agent with acting-time bonus and intrinsic reward \textbf{DDQN-Bonus} and \textbf{DDQN-Intrinsic}, respectively.

\ifx\arxiv\undefined
\textbf{Bootstrapped DDQN}
\else
\paragraph{Bootstrapped DDQN}
\fi
Another approach to improving exploration of the DQN agent is to mimic the behavior of Thompson sampling~\citep{thompson1933likelihood}.~\citet{osband2016deep} proposed the boostrapped DQN agent to achieve this goal. Here we replace the DQN loss with the DDQN loss in Eq.~\ref{eq:ddqn} and thus we name this agent bootstrapped DDQN (BootDDQN). This agent maintains an ensemble of $M$ Q-networks. For the $m$-th network, the parameters are a summation of a trainable component $\theta_m$ and a fixed randomized prior~\citep{osband2018randomized} network $\theta_m^p$, and thus the Q-network can be denoted by $Q(s, a; \widetilde{\theta}_m)$, where $\widetilde{\theta}_m:=\theta_m+\theta_m^p$. The randomized prior $\theta_m^p$ is independently initialized at the beginning of the algorithm and kept fixed during training.

During the learning process, we use the data from the replay buffer to update all the ensemble members.
\ifx\arxiv\undefined
\else
\if\arxiv1
This means that we minimize
\begin{equation}\label{eq:boot_ddqn}
L_{\text{Boot}}(\theta_1,\ldots, \theta_M)=\EE_{\cR}\frac{1}{M}\sum_{m=1}^M \big(Q(s_t, a_t; \widetilde{\theta}_m) - r_t - \gamma Q(s_{t+1}, \arg\max_{a\in\cA} Q(s_{t+1}, a;\widetilde{\theta}_m); \widetilde{\theta}'_m)
\big)^2,
\end{equation}
where $\widetilde{\theta}'_m$ is the parameter for the target network of the $m$-th ensemble member.
\fi\fi
As for acting, at the beginning of each data collection iteration, we first sample an ensemble index $m\sim\text{Unif}[M]$ and then use this ensemble member throughout this iteration, i.e., $\textsc{Act}(s) = \arg\max_{a\in\cA} Q(s, a; \widetilde{\theta}_m)$.

\ifx\arxiv\undefined
\textbf{Distributional DDQN}
\else
\paragraph{Distributional DDQN}
\fi
This agent was originally proposed by~\citet{bellemare2017distributional}. Instead of predicting the expectation of the cumulative reward using the Q-network, we predict its distribution. Define the $n$ \emph{atoms} as $v_{\min}, v_{\min} + \delta, \ldots, v_{\max}:=v_{\min}+(n-1)\delta$ and let $Q(s, a; \theta)\in\R^n$ be the PMF of a discrete distribution over the $N$ atoms and $\overline{Q}(s, a; \theta)$ be its expectation. The loss during training $L_{\text{Dist}}(\theta)$ measures the Kullback–Leibler divergence between the predicted distribution and its TD target. See~\citet{bellemare2017distributional} for more details.

\ifx\arxiv\undefined
\textbf{Policy Iteration}
\else
\paragraph{Policy Iteration}
\fi
We also experiment with an agent based on approximate policy iteration (PI). Here, we update the Q-function using least-squares Monte Carlo. We store transitions in the replay buffer in the format of $(s_t, a_t, g_t)$, where $g_t = \sum_{k=t}^{H} \gamma^{k-t} r_k$ is the empirical return, and minimize the following loss by taking a gradient step:
\begin{align}\label{eq:pi}
\ifx\arxiv\undefined
\vspace{-0.1in}
\fi
    L_{\text{PI}}(\theta) = \EE_{\cR}\left(Q(s_t, a_t; \theta) - g_t \right)^2.
\ifx\arxiv\undefined
\vspace{-0.1in}
\fi
\end{align}
Note that in standard PI, Q-functions are estimated using \emph{on-policy} data, i.e. the data generated since the most recent policy update. In this implementation, we simply sample transitions uniformly from the history, including off-policy data. This approach has been shown to implicitly regularize policy iteration updates \citep{lazic2021improved}. When acting, the agent acts either greedily with respect to $Q(s, a; \theta)$ or using an acting-time bonus as in Eq.~\ref{eq:acting_time_bonus} (PI-Bonus).

\subsection{Uncertainty Estimation}
We consider several methods for evaluating agent uncertainty: (1) standard deviation of ensemble predictions (for bootstrapped DDQN), (2) covariance of random state-action features, (3) approximate counts, and (4) random network distillation (RND) \citep{burda2018exploration}.

\ifx\arxiv\undefined
\textbf{Standard deviation of ensemble predictions}
\else
\paragraph{Standard deviation of ensemble predictions}
\fi
For Bootstrapped DDQN, we estimate the agent's uncertainty about a state-action pair $(s, a)$ as the standard deviation of the ensemble of Q-function estimates:
\ifx\arxiv\undefined
$u_{\text{std}}(s, a) = (\text{Var}\{Q(s, a; \widetilde{\theta}_m)\}_{m=1}^M)^{1/2}$.
\else
\begin{align}\label{eq:uncertainty_std}
u_{\text{std}}(s, a) = (\text{Var}\{Q(s, a; \widetilde{\theta}_m)\}_{m=1}^M)^{1/2}.
\end{align}
\fi

\ifx\arxiv\undefined
\textbf{Covariance-based uncertainty}
\else
\paragraph{Covariance-based uncertainty}
\fi
This method assumes that we have available a function $\phi: \cS \times \cA \rightarrow \mathbb{R}^d$ that maps state-action pairs to $d$-dimensional feature vectors.
While acting, we keep track of the unnormalized covariance matrix of the feature vectors, i.e.,
\ifx\arxiv\undefined
$\Phi = \sum_{t} \phi(s_t, a_t) \phi(s_t, a_t)^\top + \lambda I$,
\else
\[
\Phi = \sum_{t} \phi(s_t, a_t) \phi(s_t, a_t)^\top + \lambda I,
\]
\fi
where $\lambda$ is a regularization parameter.
We then evaluate uncertainty for a state-action pair $(s, a)$ as:
\ifx\arxiv\undefined
$u_{\text{cov}}(s, a) = (\phi(s, a)^\top \Phi^{-1} \phi(s, a))^{1/2}$.
\else
\begin{align}\label{eq:uncertainty_cov}
 u_{\text{cov}}(s, a) = (\phi(s, a)^\top \Phi^{-1} \phi(s, a))^{1/2}.
\end{align}
\fi

We can extract the feature $\phi$ using a pre-trained representation, a randomly initialized neural network, or a combination of both. In this work, we extract random Fourier features \citep{rahimi2007random} from the state, denoted as $\psi(s)$ and compute state-action features as $\phi(s, a) = \psi(s) \otimes e_a$, where $e_a$ is an $|\cA|$-dimensional action indicator vector. Note that in practice, we can maintain the matrix $\Phi^{-1}$ in a computationally efficient manner by leveraging the Sherman–Morrison formula~\citep{sherman1950adjustment}.

\ifx\arxiv\undefined
\textbf{Approximate counts}
\else
\paragraph{Approximate counts}
\fi
Another uncertainty metric is to keep approximate counts of the state-action pairs. More specifically, we design a discretization of the state space $\cS$, denoted by $\overline\cS$ ($|\overline{\cS}|<\infty$). Let $\psi:\cS\mapsto\overline{\cS}$ be the function that maps a state to its corresponding discrete element in $\overline{\cS}$. Then we can use the following uncertainty metric
\ifx\arxiv\undefined
$u_{\text{count}}(s, a) = (n(s, a)+\lambda)^{-1/2}$,
\else
\begin{align}\label{eq:uncertainty_count}
  u_{\text{count}}(s, a) = (n(s, a)+\lambda)^{-1/2},
\end{align}
\fi
where $n(s, a)$ is the visitation count of $(\psi(s), a)$ in the $\overline{\cS}\times\cA$ space. Note that the approximate-count based method is a special case of the covariance-based uncertainty with $\psi(\cdot)$ considered as a one-hot encoded feature vector.

In this paper, we use this method particularly for image observations in Atari games. More specifically, we downsample the image to a smaller size by average pooling, and then discretize the pixel values. Using the terminology in Go-Explore~\citep{ecoffet2019go}, we call each discrete element in $\overline{\cS}$ a \emph{cell}. Although Go-Explore uses a similar downsampling method, the state-revisiting rule in our UFLP framework is much simpler than in Go-Explore.

\ifx\arxiv\undefined
\textbf{Random network distillation (RND)}
\else
\paragraph{Random network distillation (RND)}
\fi
In RND, uncertainty is given by the error of a neural network $\widehat{f}:\cS\mapsto\R^k$ trained to predict the features of the observations given by a fixed randomly initialized neural network $f:\cS\mapsto\R^k$.
The $\ell_2$ error is used as the uncertainty metric for the states, i.e.,
\ifx\arxiv\undefined
$u_{\text{rnd}}(s) = \|\widehat{f}(s) - f(s)\|^2$.
\else
\begin{align}\label{eq:uncertainty_rnd}
  u_{\text{rnd}}(s) = \|\widehat{f}(s) - f(s)\|^2.
\end{align}
\fi
The use of this metric in our work differs from the original RND work of \citet{burda2018exploration}, where this error is used as an intrinsic reward.
\ifx\arxiv\undefined
\vspace{-0.1in}
\fi
\section{Experiments}
\ifx\arxiv\undefined
\vspace{-0.05in}
\fi

In this section, we evaluate the benefits of local vs. online access by training agents on difficult exploration tasks. We use two (\texttt{bsuite}) \citep{osband2019behaviour} environments: Deep Sea and Cartpole Swingup,  and four Atari games: Montezuma's Revenge, PrivateEye,  Venture, and Pitfall. These games are known to correspond to difficult exploration problems \citep{badia2020agent57}. In Appendix~\ref{sec:checkpointing}, we provide details on how to checkpoint and restore the environment state using Python, for both \bsuite~and Atari. We also provide our hyperparameter choices in Appendix~\ref{sec:hprams}. For all the figures in this section, the shaded area shows the $95\%$ confidence interval.

\subsection{Behavior Suite Experiments}\label{sec:bsuite}

\begin{figure}[ht]
\centering
\subfigure[Deep Sea]{\includegraphics[width=.35\linewidth]{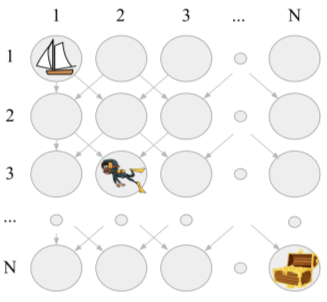}}
\subfigure[Cartpole Swingup]{\includegraphics[width=.35\linewidth]{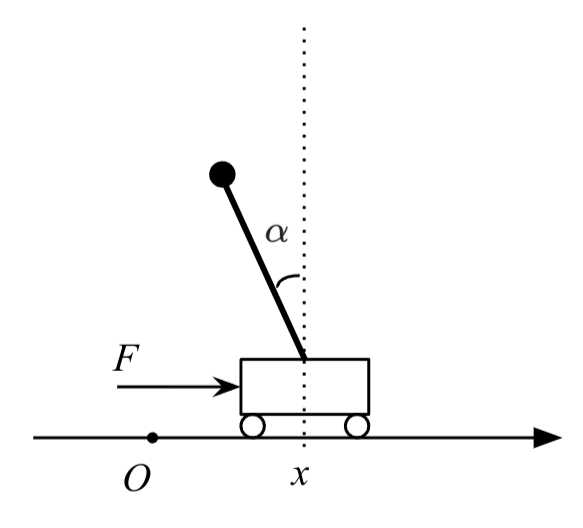}}
\ifx\arxiv\undefined
\vspace{-0.1in}
\fi
\caption{\bsuite~environments}
\ifx\arxiv\undefined
\vspace{-0.15in}
\fi
\label{fig:bsuite}
\end{figure}

We first introduce the two \texttt{bsuite} environments that we use in this section. Illustrations of the two environments can be found in Figure~\ref{fig:bsuite}.

\ifx\arxiv\undefined
{\bf Deep Sea\;}
\else
\paragraph{Deep Sea}
\fi
The environment is an $N \times N$ grid with one-hot state encoding. The agent starts from the top left corner of the grid and descends down-left or down-right in each timestep, depending on the action. There is a small cost of $r = - 0.01/N$ for moving right, and $r=0$ for moving left. If the agent reaches the bottom-right corner, taking only ``right'' actions, it gets a reward of $+1$. This is a simple but challenging exploration problem, due to the fact that exploring uniformly at random only has a $2^{-N}$ chance of finding the high-reward state.

\ifx\arxiv\undefined
{\bf Cartpole Swingup \;}
\else
\paragraph{Cartpole Swingup}
\fi
The goal is to swing up and balance an unactuated pole by applying forces to a cart at its base. The physics model conforms to \citet{barto1983neuronlike}. The pole starts from a random position pointing down, and the agent can apply a force of $-1$, $0$, or $1$ to the cart. There is a small cost of $r=-0.1$ for applying a non-zero force. The agent gets a reward of $+1$ if the pole is within a small angle of being upright and the cart is within a small range around the origin. We consider two versions of this environment: \emph{default} and \emph{hard} versions, where the hard version has sparser rewards. More details on how the rewards are defined in the two versions can be found in Appendix~\ref{sec:hparams_cartpole}.

Our experiments for \bsuite~environments involve four agents: BootDDQN, DDQN-Bonus, vanilla DDQN, and PI-Bonus. For BootDDQN, we use standard deviation of the ensemble as the uncertainty metric $u_{\text{std}}$. For all other agents, we use the covariance-based uncertainty $u_{\text{cov}}$. The confidence intervals are calculated with $10$ random seeds for Deep Sea and $20$ seeds for Cartpole Swingup. DDQN-Intrinsic does not converge in our Deep Sea and Cartpole Swingup experiments due to numerical issues. Therefore, we do not report the results of DDQN-Intrinsic in this section.

\begin{figure}[ht]
\centering
\ifx\arxiv\undefined
\includegraphics[width=.9\linewidth]{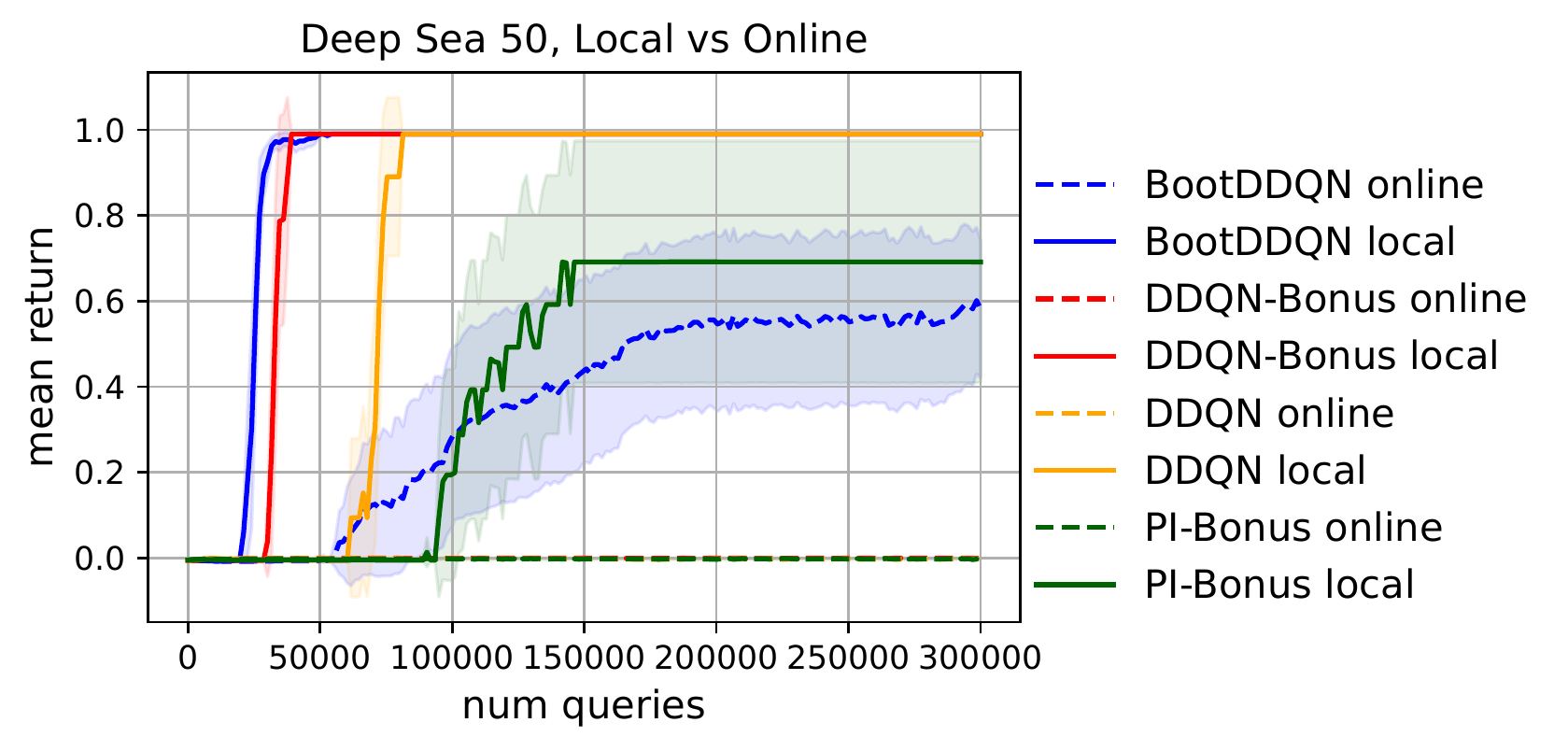}
\vspace{-0.1in}
\else
\includegraphics[width=.7\linewidth]{figures/deep_sea_50.pdf}
\fi
\caption{Local and online access on Deep Sea 50. Curves with the same color correspond to the same agent. Dashed and solid curves correspond to online and local access, respectively.}
\label{fig:deep_sea_50}
\end{figure}

\begin{figure}[ht]
\centering
\ifx\arxiv\undefined
\subfigure[The effect of $\pinit$]{\includegraphics[width=.47\linewidth]{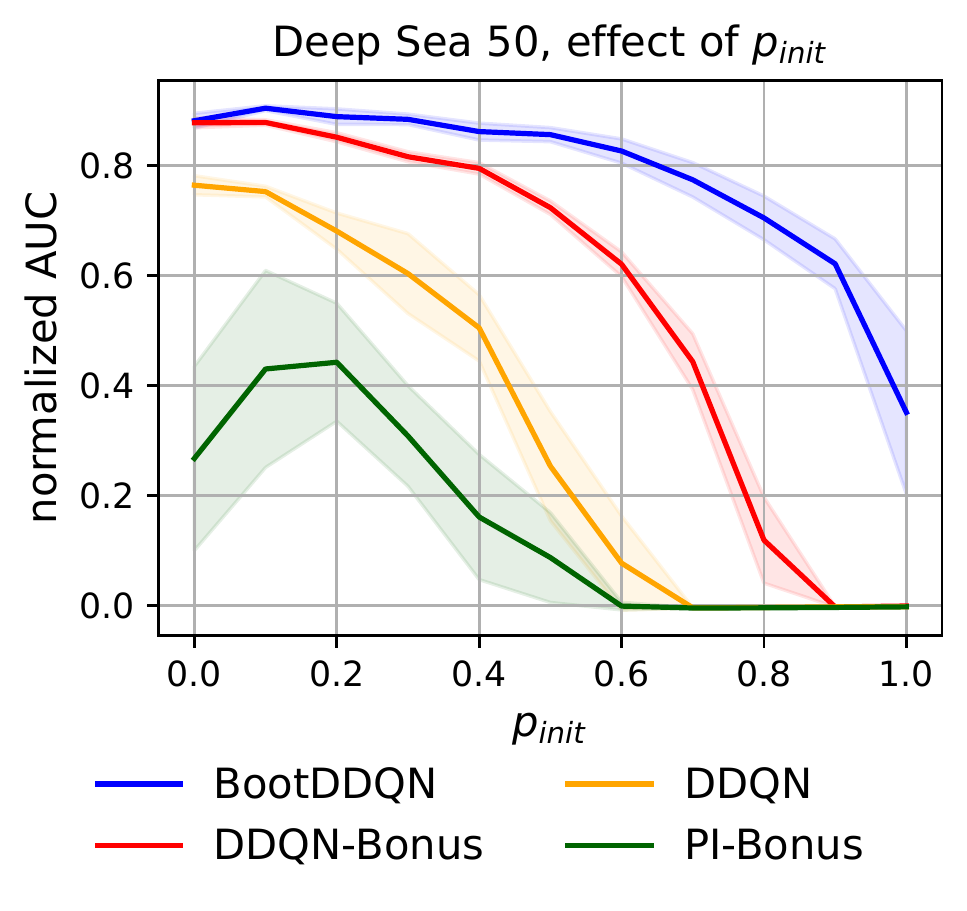}}
\subfigure[The effect of $B$]{\includegraphics[width=.48\linewidth]{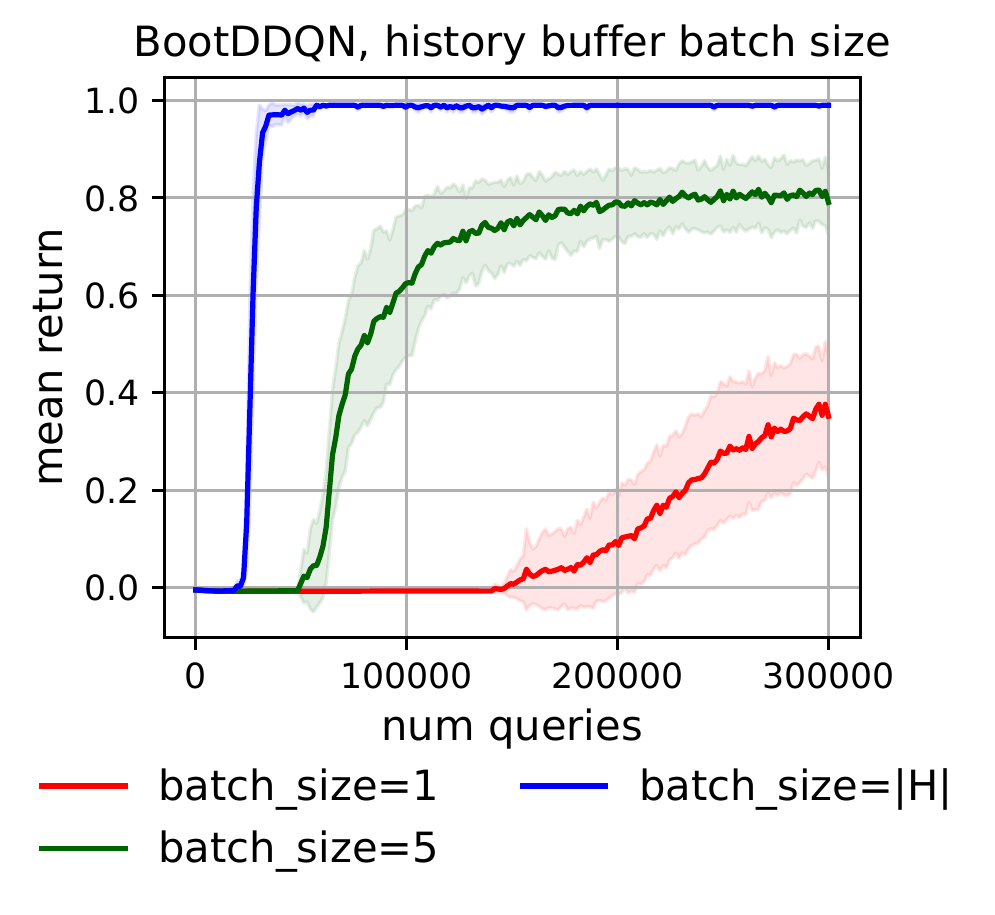}}
\vspace{-0.1in}
\else
\subfigure[The effect of $\pinit$]{\includegraphics[width=.37\linewidth]{figures/deep_sea_50_pinit.pdf}}
\subfigure[The effect of $B$]{\includegraphics[width=.38\linewidth]{figures/deep_sea_50_boot_batch_size.pdf}}
\fi
\caption{The effect of $\pinit$ and history buffer batch size $B$ in Deep Sea 50. Choosing $B=1$ is equivalent to choosing a random element in the history buffer, without considering uncertainty.}
\label{fig:deep_sea_ablation}
\end{figure}

\begin{figure}[ht]
\centering
\ifx\arxiv\undefined
\vspace{-0.1in}
\includegraphics[width=.55\linewidth]{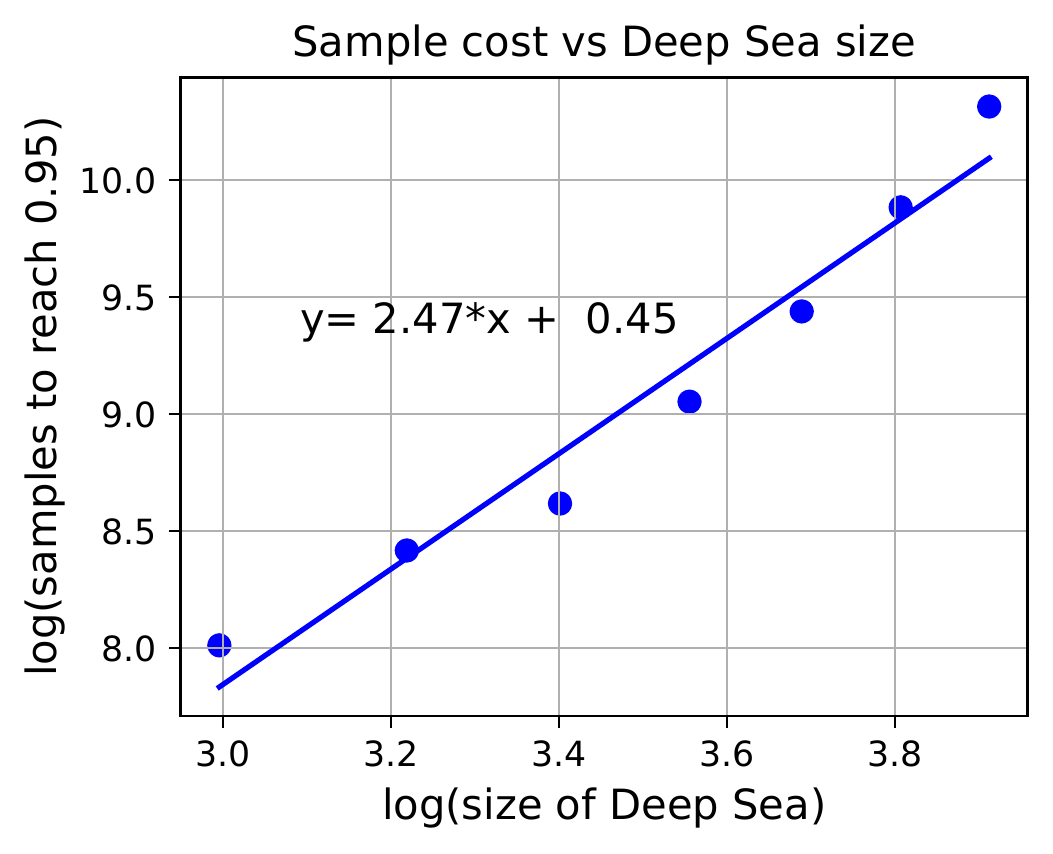}
\vspace{-0.1in}
\else
\includegraphics[width=.4\linewidth]{figures/deep_sea_scaling.pdf}
\fi
\caption{The number of queries needed to achieve mean return $0.95$ vs the size of the Deep Sea environment (BootDDQN).}
\ifx\arxiv\undefined
\vspace{-0.2in}
\fi
\label{fig:deep_sea_scaling}
\end{figure}

\begin{figure*}[ht]
\centering
\subfigure[]{\includegraphics[width=.24\linewidth]{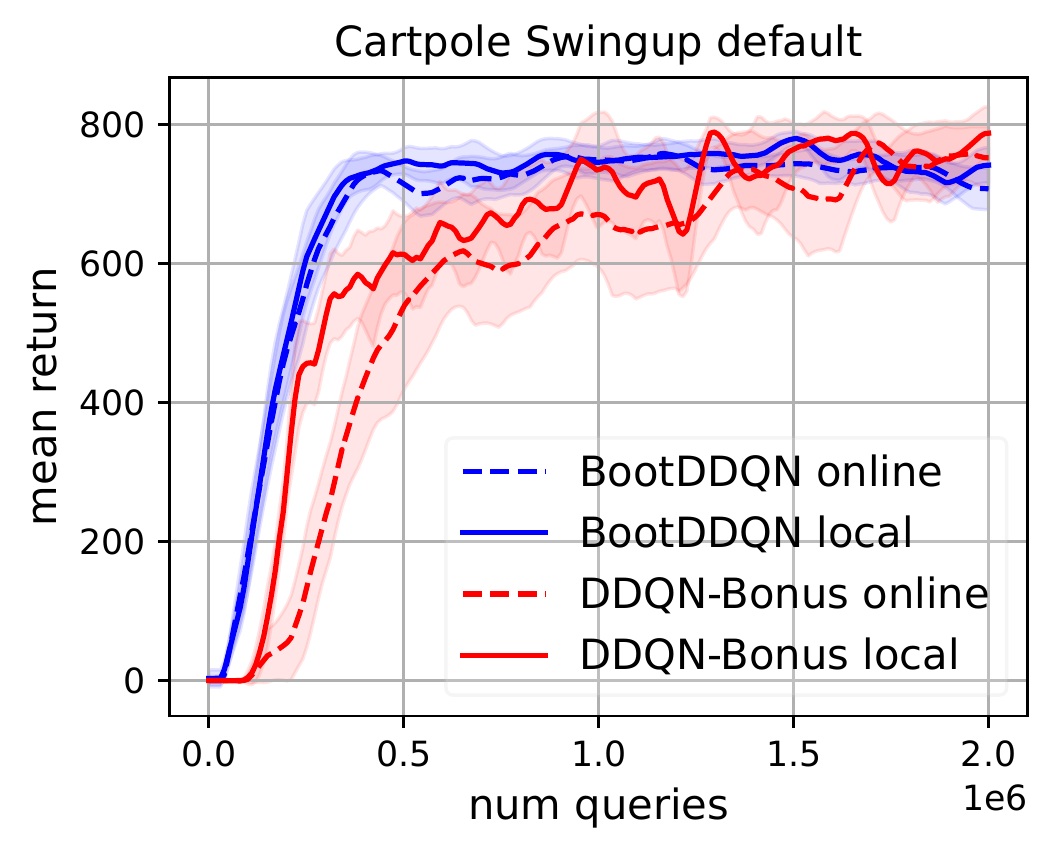}}
\subfigure[]{\includegraphics[width=.24\linewidth]{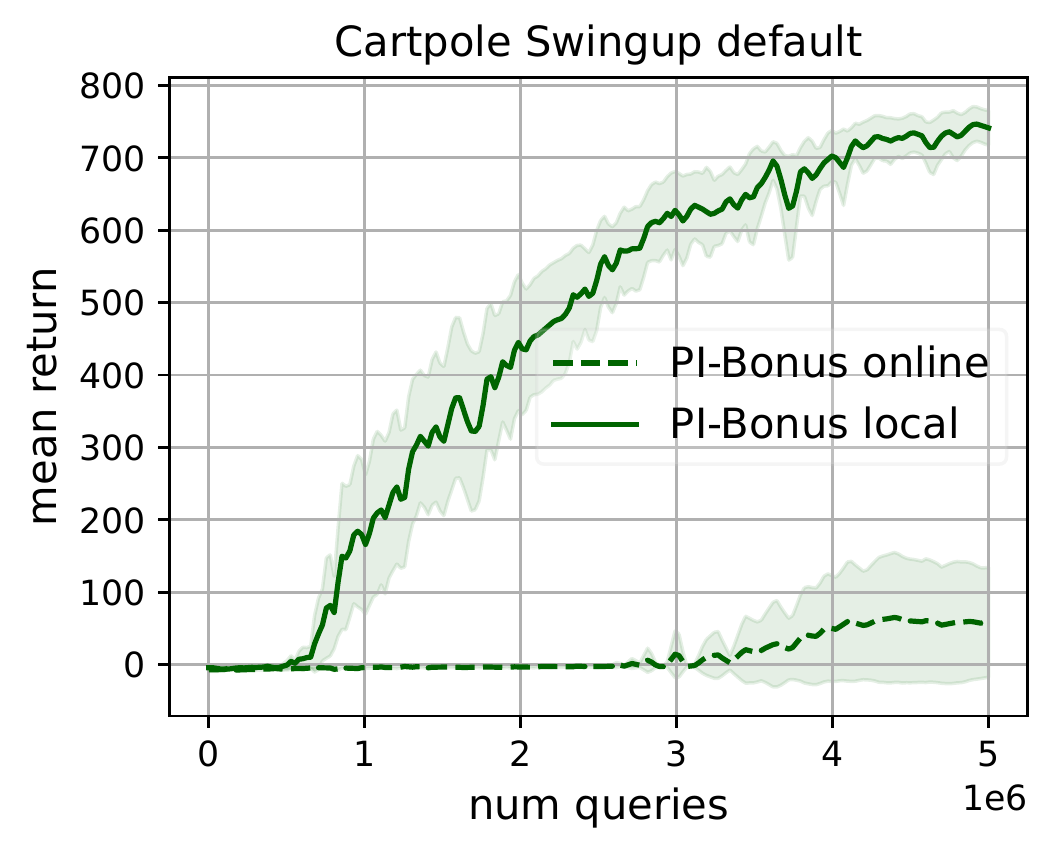}}
\subfigure[]{\includegraphics[width=.25\linewidth]{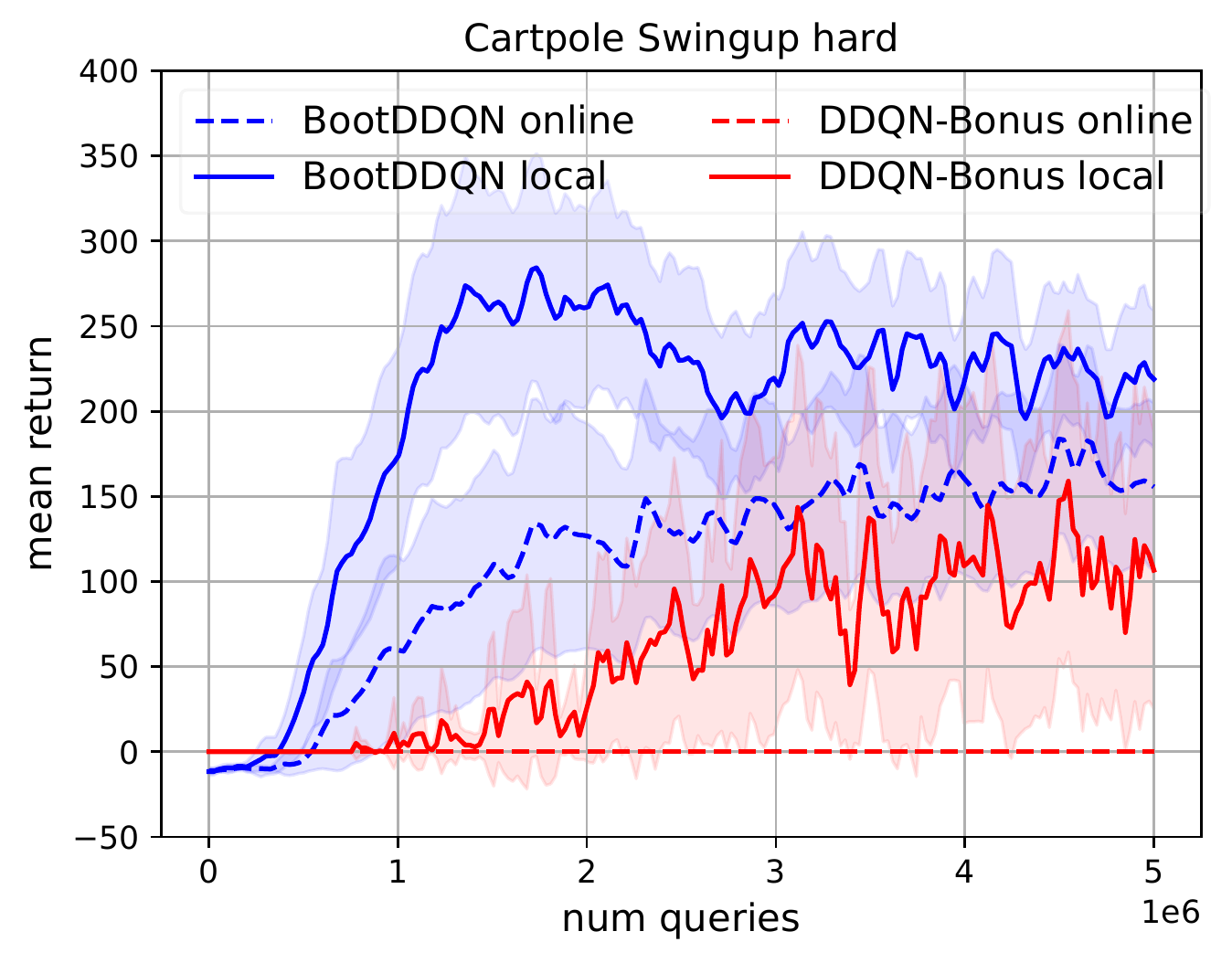}}
\subfigure[]{\includegraphics[width=.24\linewidth]{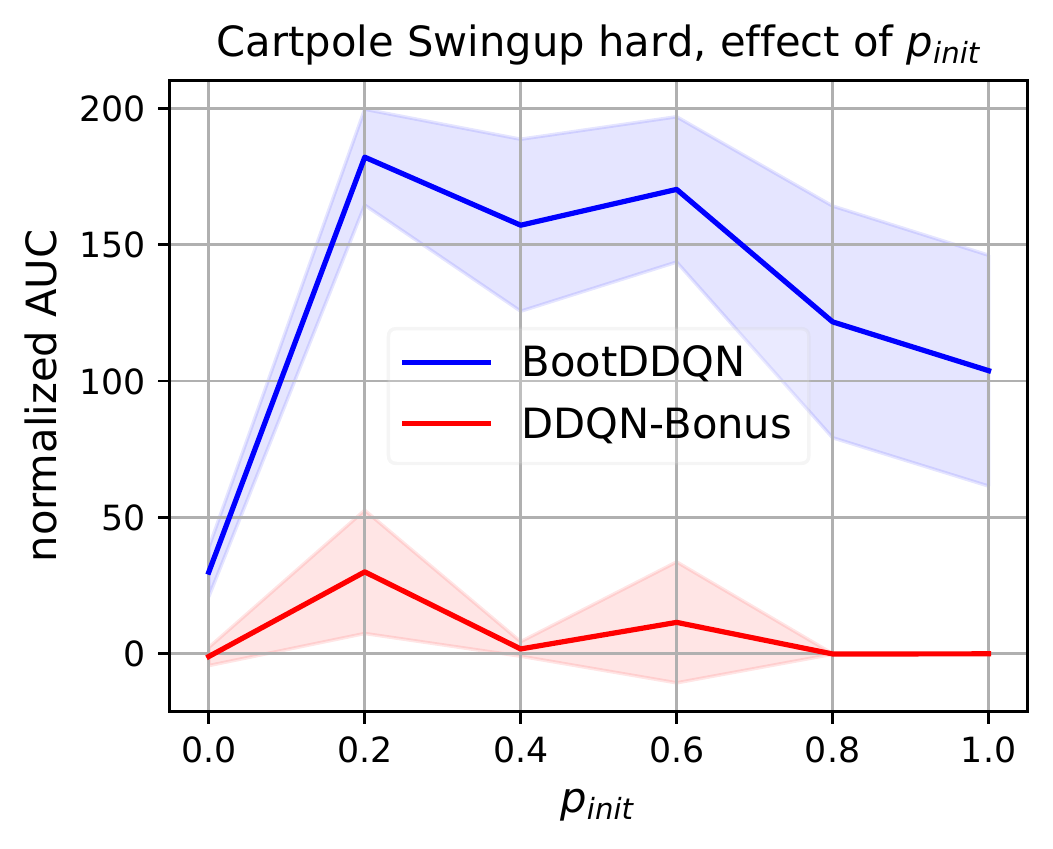}}
\ifx\arxiv\undefined
\vspace{-0.1in}
\fi
\caption{Cartpole Swingup results. (a) BootDDQN and DDQN-Bonus in the default version. (b) PI-Bonus in the default version. (c) BootDDQN and DDQN-Bonus in the hard version. (d) The effect of $\pinit$ in the hard version for BootDDQN and DDQN-Bonus.}
\ifx\arxiv\undefined
\vspace{-0.1in}
\fi
\label{fig:cartpole_swingup}
\end{figure*}

\ifx\arxiv\undefined
\textbf{Deep Sea results}
\else
\paragraph{Deep Sea results}
\fi
We first compare the return of the agents on Deep Sea with size $N=50$ in the online and local access (UFLP) settings. We use $\pinit=0.1$ and history buffer batch size $B=|\cH|$ (i.e., we choose the most uncertain state-action pair from the entire history buffer) for all the local access runs. As we can see from Figure~\ref{fig:deep_sea_50}, local access leads to significantly higher mean return for each agent. In fact, in the online setting, except for BootDDQN, none of the agents can get a return that is significantly higher than $0$ within $3\times 10^5$ simulator queries. Therefore, in this hard-exploration environment, local access significantly improves the sample efficiency.

We also investigate the role of $\pinit$ and history buffer batch size $B$ in Deep Sea. In Figure~\ref{fig:deep_sea_ablation}(a), we plot the normalized area under curve (AUC) for the the convergence curves in Figure~\ref{fig:deep_sea_50} as a function of $\pinit$.\footnote{By normalization, we mean that we divide the AUC by the total number of simulator queries during training.} This quantity reflects how fast the return converges. The best performance  is achieved with a relatively small $\pinit$ (e.g., $0.0\sim 0.3$) for most agents. This demonstrates the benefits of data collection from intermediate states.  In Figure~\ref{fig:deep_sea_ablation}, we compare the sample efficiency of BootDDQN with $B=1, 5,$ and $|\cH|$. As we can see, the best performance is achieved with $B=|\cH|$, i.e., choosing the most uncertain element in the history buffer. This demonstrates the importance of starting from an uncertain state in Deep Sea. In Appendix~\ref{sec:additiona_exp_deep_sea}, we provide similar results for other agents.

In Figure~\ref{fig:deep_sea_scaling}, we show the number of queries needed to achieve mean return $0.95$, for the BootDDQN agent, as a function of the size of the Deep Sea environment $N$, to illustrate the scaling of the sample cost. Our results indicates a near-optimal dependency of $\cO(N^{2.47})$. \footnote{The optimal sample scale is $\cO(N^2)$, proportional to the number of all possible state-action pairs.}

\ifx\arxiv\undefined
\textbf{Cartpole Swingup results}
\else
\paragraph{Cartpole Swingup results}
\fi
For the default version of Cartpole Swingup, we find that for BootDDQN and DDQN-Bonus agents, the sample efficiency of online and local access modes are similar (Figure~\ref{fig:cartpole_swingup}(a)), whereas for the PI-Bonus agent, local access leads to a significant improvement (Figure~\ref{fig:cartpole_swingup}(b)). This indicates that for environments with relatively dense reward, the benefit of local access can be small, especially for value-based agents. For the hard version, for both BootDDQN and DDQN-Bonus agents, we find that local access leads to a significant improvement over online access (Figure~\ref{fig:cartpole_swingup}(c)). We did not observe positive rewards in the hard version of Cartpole Swingup using the PI-Bonus agent, regardless of the access protocol. In Figure~\ref{fig:cartpole_swingup}(d), we  show that the best performance can be achieved with a relatively small $\pinit$, e.g., $0.2$, for the hard version of Cartpole Swingup. One observation is that when $\pinit=0.0$, the performance is bad. We hypothesize that this is because for $\pinit=0.0$, we only observe a single initial state from the initial-state distribution, and the agent may not have enough information on how to act from other initial states. 

\ifx\arxiv\undefined
\vspace{-0.05in}
\fi
\subsection{Atari}\label{sec:atari}
\ifx\arxiv\undefined
\vspace{-0.05in}
\fi

In this section, we evaluate our approach on four Atari games from the Arcade Learning Environment (ALE) \citep{bellemare2013arcade}:

\begin{figure}[ht]
\centering
\ifx\arxiv\undefined
\includegraphics[width=.95\linewidth]{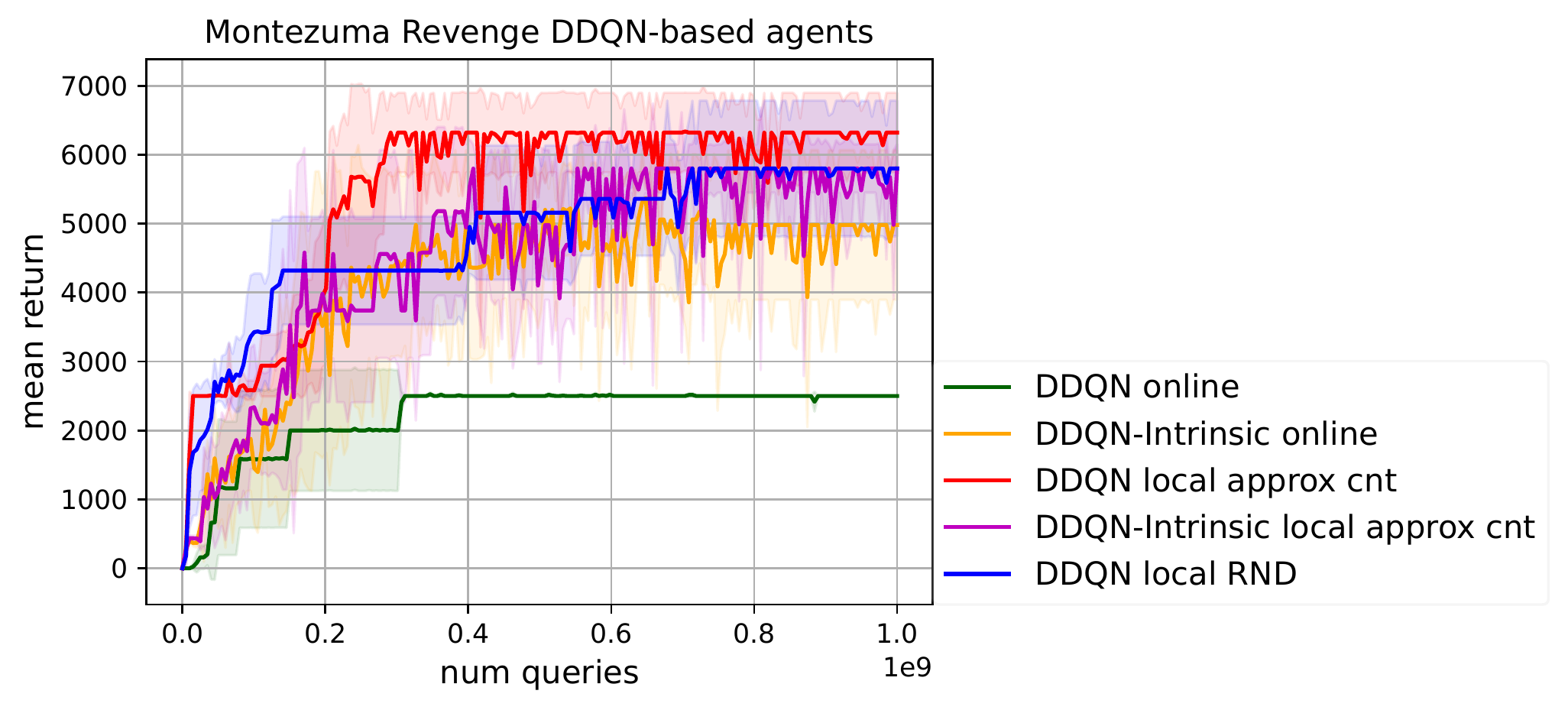}
\vspace{-0.1in}
\else
\includegraphics[width=.7\linewidth]{figures/montezuma_ddqn.pdf}
\fi
\caption{Uncertainty-first local planning vs. online access on Montezuma's Revenge with DDQN-based agents.}
\ifx\arxiv\undefined
\vspace{-0.05in}
\fi
\label{fig:montezuma_ddqn}
\end{figure}

\begin{figure}[ht]
\centering
\ifx\arxiv\undefined
\subfigure[]{\includegraphics[width=.47\linewidth]{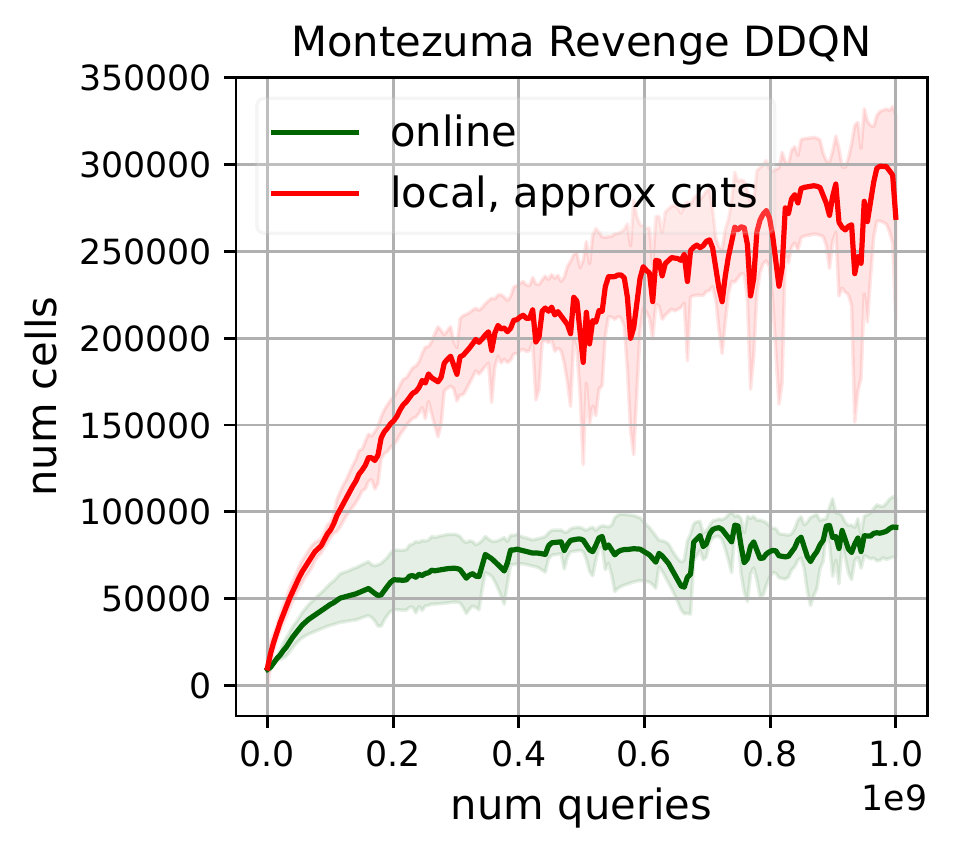}}
\subfigure[]{\includegraphics[width=.49\linewidth]{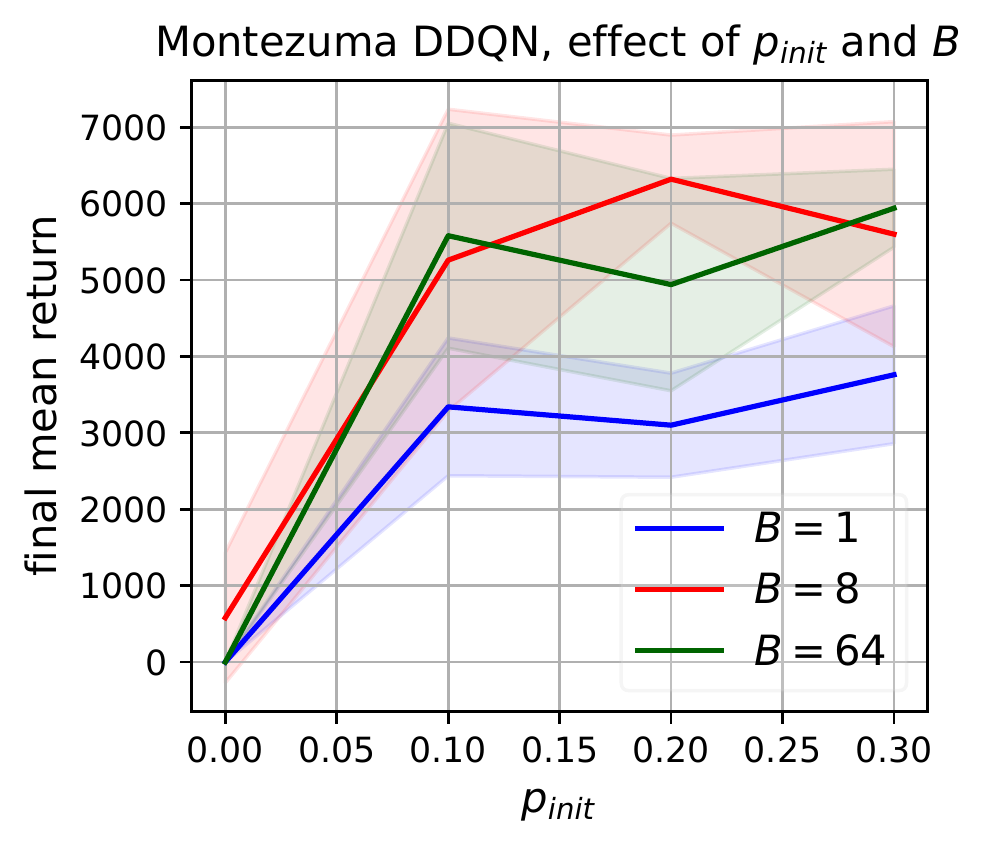}}
\vspace{-0.1in}
\else
\subfigure[]{\includegraphics[width=.37\linewidth]{figures/montezuma_num_cells.pdf}}
\subfigure[]{\includegraphics[width=.39\linewidth]{figures/montezuma_pinit_and_b.pdf}}
\fi
\caption{(a) The number of cells found by the DDQN agent in the online and local settings using the approximate-count-based uncertainty. (b) The effect of $\pinit$ and history buffer batch size $B$. Again we note that choosing $B=1$ is equivalent to choosing a random element in the history buffer, without considering uncertainty.}
\ifx\arxiv\undefined
\vspace{-0.1in}
\fi
\label{fig:montezuma_ablation}
\end{figure}

\begin{figure*}[ht]
\centering
\includegraphics[width=\linewidth,trim={100 5 100 10},clip]{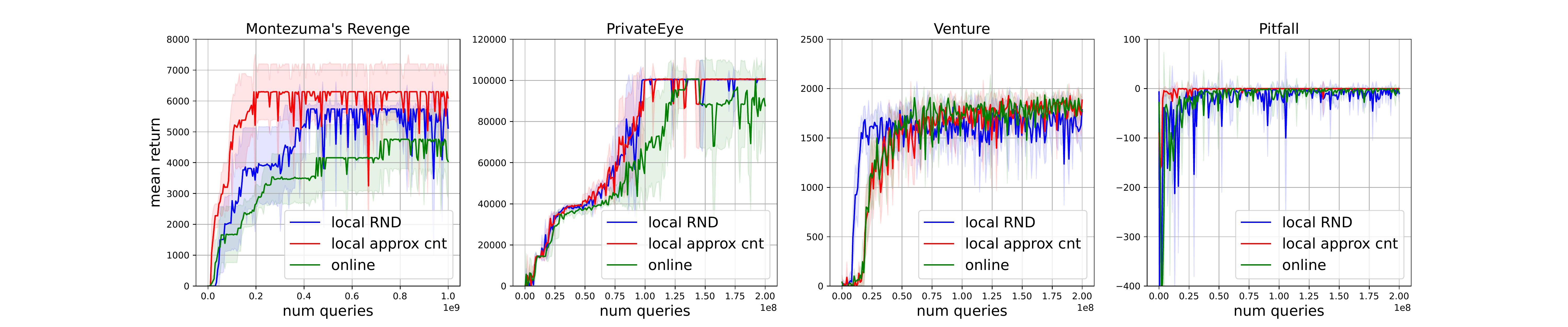}
\ifx\arxiv\undefined
\vspace{-0.3in}
\fi
\caption{Uncertainty-first local planning vs. online access on hard-exploration Atari games with a distributional DDQN agent.}
\ifx\arxiv\undefined
\vspace{-0.15in}
\fi
\label{fig:distddqn}
\end{figure*}

\ifx\arxiv\undefined
{\bf Montezuma's Revenge \;}
\else
\paragraph{Montezuma's Revenge}
\fi
Montezuma's Revenge (MR) has sparser rewards than most ALE environments. The agent only receives positive rewards after performing a long series of specific actions.
MR has been viewed as one of the most difficult exploration challenges for deep RL. A few approaches surpassing average human performance of 4753 points \citep{badia2020agent57} include RND~\citep{burda2018exploration}, NGU~\citep{badia2020never}, Agent57~\citep{badia2020agent57}, and MEME~\citep{kapturowski2022human}.
The SOTA of 43K is achieved by Go-Explore~\citep{ecoffet2021first}.

\ifx\arxiv\undefined
{\bf Pitfall \;}
\else
\paragraph{Pitfall}
\fi
This is another highly challenging exploration game in ALE. In addition to sparse positive rewards, Pitfall includes distractor rewards, such as small negative rewards for hitting an enemy, and is only partially observable. Most agents obtain zero reward, and a few exceptions include NGU, Agent57, MEME, and Go-Explore.

\ifx\arxiv\undefined
{\bf PrivateEye \;}
\else
\paragraph{PrivateEye}
\fi
This is also a sparse reward game. The average human performance is 69K~\citep{badia2020agent57}.

\ifx\arxiv\undefined
{\bf Venture \;}
\else
\paragraph{Venture}
\fi
Rewards are denser than in the $3$ other games, and the average human score is 1187 \citep{badia2020agent57}.

As base agents, we use the DDQN and distributional DDQN implementations in the Acme framework \citep{hoffman2020acme} for distributed RL. In Acme, agent functionality is split into multiple actors, which collect data and write to the replay buffer, and a single learner which updates the agent parameters based on replay data. In our implementation, the actors also write to a common (smaller) history buffer, and choose states to reset to from the history buffer based on uncertainty. 
We experiment with two different uncertainty metrics: approximate counts and RND. For approximate counts, we downsample the grayscale game images to a shape of $(12, 12)$, and discretize the pixel values to $8$ levels. For RND, the random features are extracted by the Q-network architecture followed by a 2-layer MLP with embedding size $1024$. The predictor also has the same architecture. More implementation details and hyperparameters are given in Appendix~\ref{sec:hparams_atari}. All the confidence intervals in this section are calculated using $5$ random seeds.

We first evaluate agents on Montezuma's Revenge. For the online setting, we use the vanilla DDQN and DDQN-Intrinsic agents, with the intrinsic reward based on approximate-count uncertainty. For the local setting, we combine DDQN with both approximate-count and RND based uncertainty. We also experiment with DDQN-Intrinsic in the local setting with approximate counts. 

Figure~\ref{fig:montezuma_ddqn} shows that local access significantly improves the final return over online access for both uncertainty metrics. Here, we emphasize that reaching a higher final return using the same number of simulator queries implies that an algorithm has better sample efficiency.
In Figure~\ref{fig:montezuma_ablation}(a), we plot the number of cells (when using approximate counts) found by the online and local DDQN agents. As expected, in the local access setting, the agent finds more cells, indicating that a larger state space is discovered. Similar results for DDQN-Intrinsic can be found in Appendix~\ref{sec:additiona_exp_atari}. In Figure~\ref{fig:montezuma_ablation}(b), we  study the role of $\pinit$ and history buffer batch size $B$ in DDQN with approximate counts. We can see that $B=8, 64$ is significantly better than $B=1$, indicating the importance of starting from an uncertain state. We also observe that $\pinit=0.0$ is a bad choice in this game. We hypothesize that this is due to the fact that in our implementation, the actors only maintain the approximate counts of the cells that they have visited, rather than cells that exist in the replay buffer. Thus, when $\pinit=0.0$, there are not enough data in the replay buffer that correspond to the state space near the initial state, and as the result the agent forgets how to act at the beginning of the episode. However, choosing $\pinit\in[0.1, 0.3]$ leads to similar performance.

We also observe improvement for the DDQN-Intrinsic agent using local access (Figure~\ref{fig:montezuma_ddqn}). However, since DDQN-Intrinsic already has an exploration bonus in the intrinsic reward, the additional benefit of local planning is relatively small. We note that for DDQN-Intrinsic, to get good performance in the local setting, we need to choose a larger value of $\pinit$, i.e., $0.7$. This means that we need to reduce the amount of local access iterations in order to reach a good balance between exploration and exploitation.

We evaluate the performance of distributional DDQN with UFLP ($\pinit=0.3$) for both approximate-count and RND uncertainty on all four Atari games. 
As we can see in Figure~\ref{fig:distddqn}, on Montezuma's Revenge, local planning dramatically improves the score of the baseline algorithm to a super-human level. On PrivateEye, local access improves the sample complexity and stability of the baseline algorithm. Venture results are neutral, possibly because the rewards are relatively dense and thus the exploration problem is less challenging compared to other games. On Pitfall, both local and and online access versions fail to obtain positive scores. We conjecture that this is due to the partially-observable nature of this MDP; indeed, prior works that have obtained positive scores have relied on side information, stateful policies, or both \citep{badia2020agent57,ecoffet2019go,ecoffet2021first}. 

We summarize all Atari results in Appendix~\ref{sec:additiona_exp_atari} and include the highest scores obtained by actors during data collection. Note that actor returns are sometimes considerably higher than those of the learned policy (greedy w.r.t. the Q-function), e.g. $14300$ vs. $7100$ for Montezuma's Revenge and $1800$ vs. $0$ for Pitfall. This suggests the performance of the DDQN agents could be improved by more effective learning (not addressed here) in addition to exploration. 

\ifx\arxiv\undefined
\vspace{-0.1in}
\fi
\section{Conclusions and Future Directions}
\ifx\arxiv\undefined
\vspace{-0.01in}
\fi

We propose a new algorithmic framework for learning  with a simulator under the local access protocol.
We demonstrate that our proposed uncertainty-first approach to revisiting states in history can dramatically improve the sample cost of several baseline algorithms on sparse-reward environments. An important direction for future work is improving the quality of uncertainty estimation in MDPs, since the this directly affects the effectiveness of the framework. Another interesting direction for future work is to extend this approach to partially observed environments.

\bibliography{reference}

\appendix
\section{Additional Experimental Results}\label{sec:additional_experiments}

\subsection{Deep Sea}\label{sec:additiona_exp_deep_sea}

\begin{figure}[ht]
\centering
\begin{subfigure}
  \centering
  \includegraphics[width=.3\linewidth]{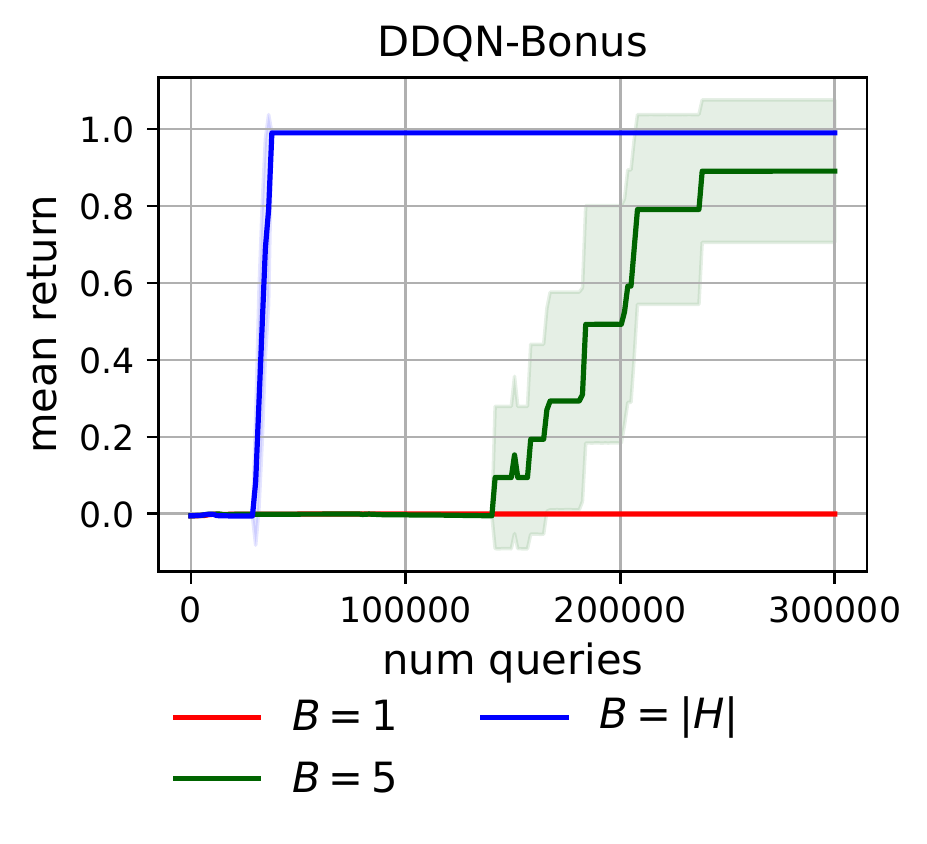}
\end{subfigure}%
\begin{subfigure}
  \centering
  \includegraphics[width=.29\linewidth]{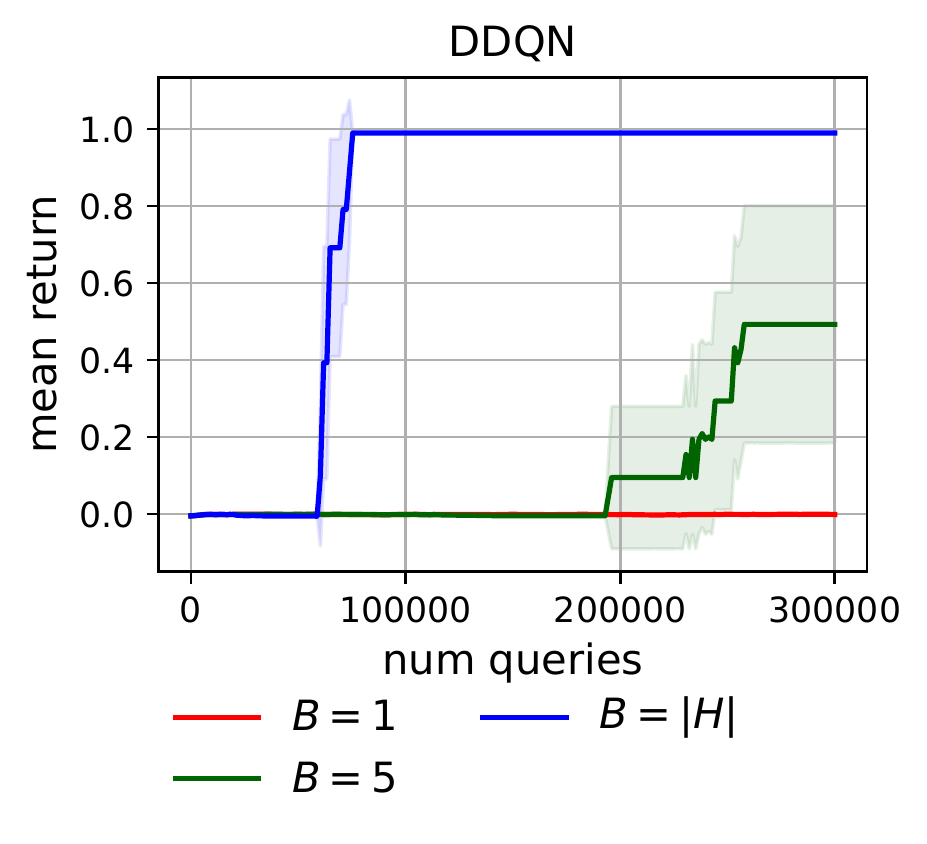}
\end{subfigure}
\begin{subfigure}
  \centering
  \includegraphics[width=.3\linewidth]{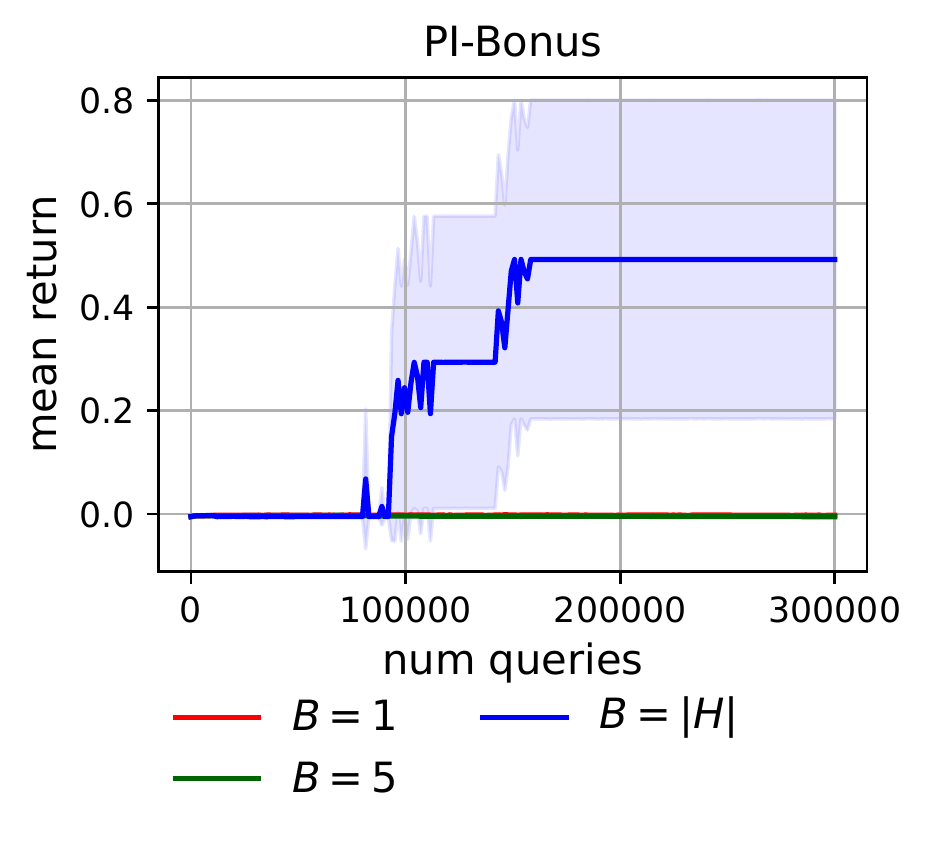}
\end{subfigure}
\caption{The effect of history buffer batch size $B$ for the DDQN-Bonus, DDQN, and PI-Bonus agents in Deep Sea 50.}
\label{fig:deep_sea_additional}
\end{figure}

In Figure~\ref{fig:deep_sea_additional}, we show the effect of history buffer batch size $B$ for the DDQN-Bonus, DDQN, and PI-Bonus agents in Deep Sea 50. As we can see, the best performance is achieved with $B=|\cH|$. Therefore, it is important to start the data collection process with an uncertain state.

\subsection{Cartpole Swingup}\label{sec:additiona_exp_cartpole}

\begin{figure}[ht]
\centering
\subfigure[]{\includegraphics[width=.3\linewidth]{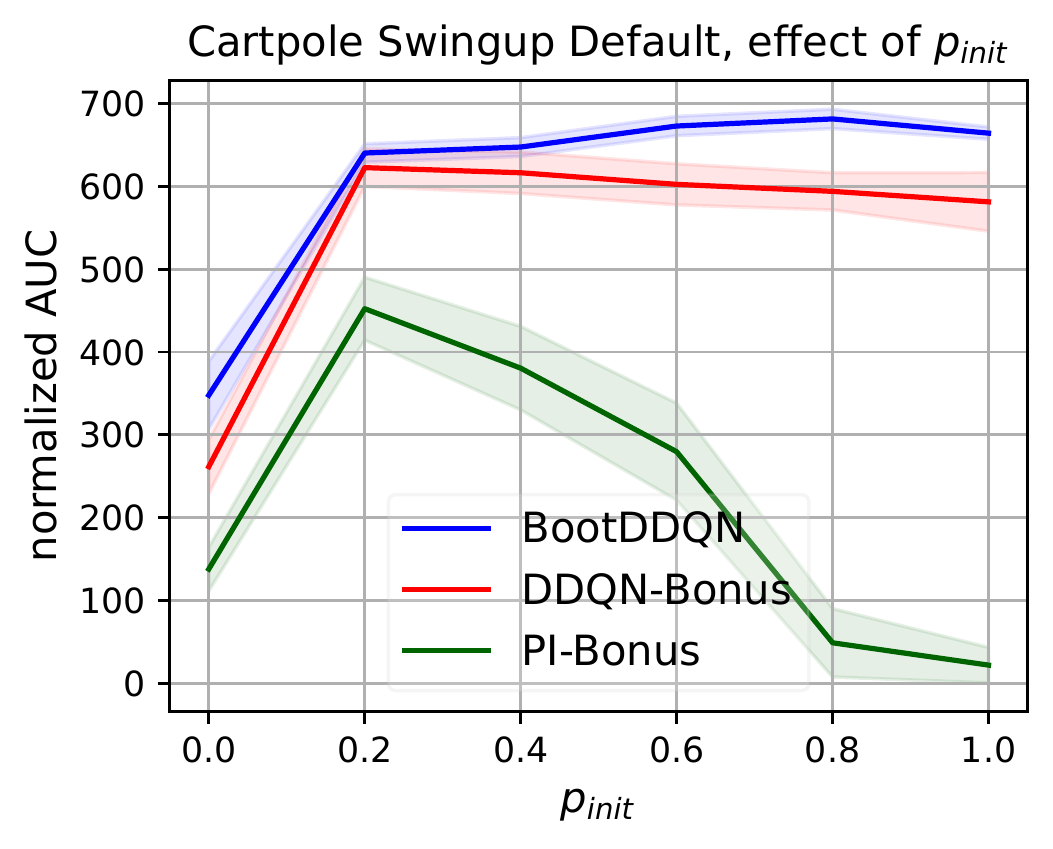}}
\subfigure[]{\includegraphics[width=.3\linewidth]{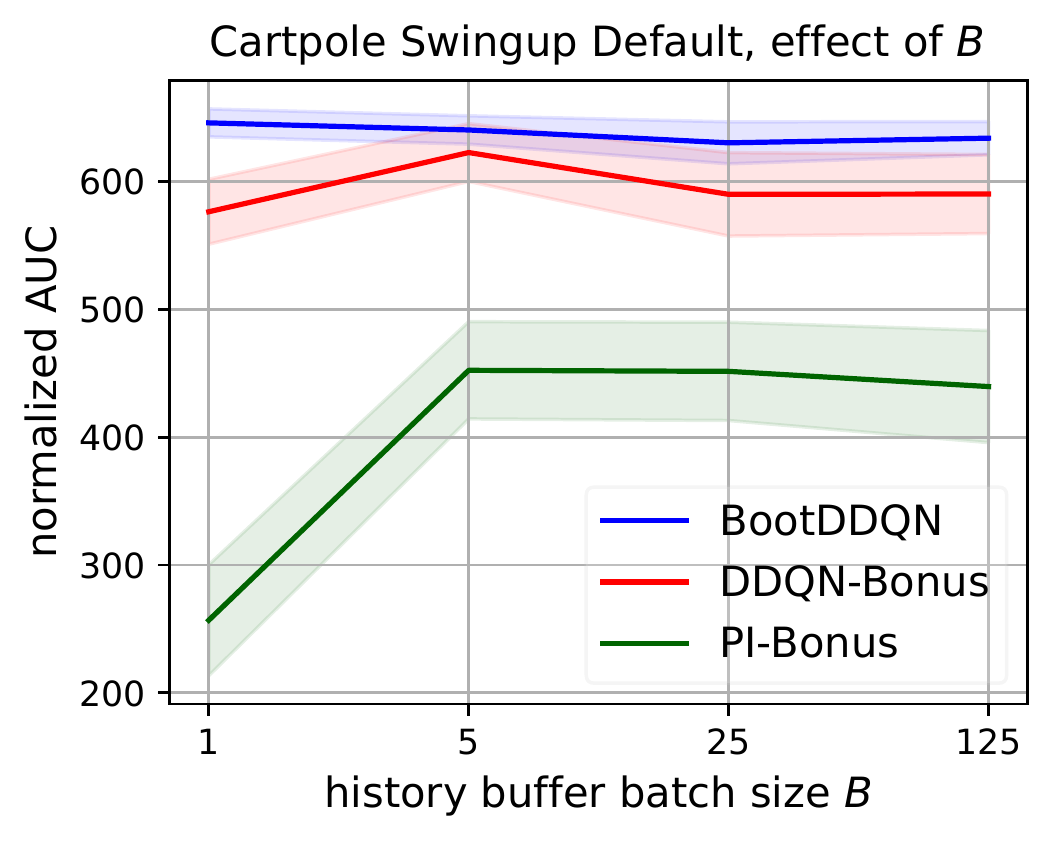}}
\subfigure[]{\includegraphics[width=.3\linewidth]{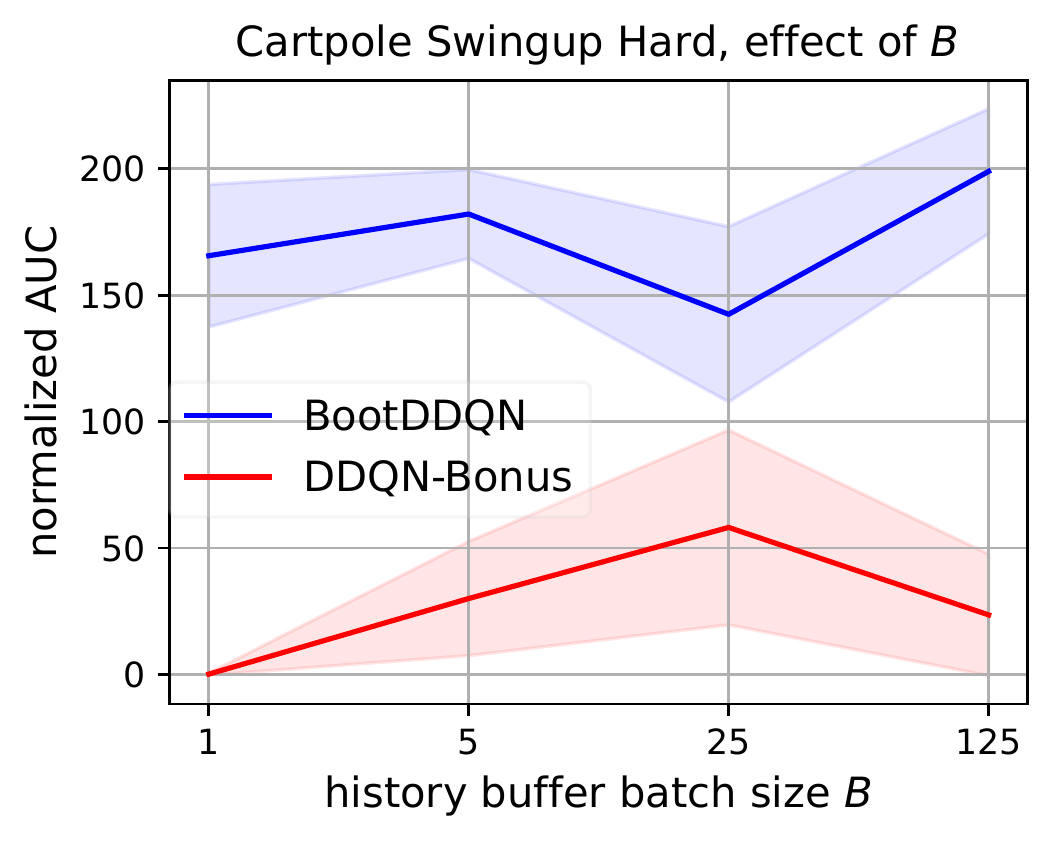}}

\caption{(a) The effect of $\pinit$ in the default version of Cartpole Swingup. (b) The effect of history buffer batch size $B$ in the default version of Cartpole Swingup. (c) The effect of history buffer batch size $B$ in the hard version of Cartpole Swingup.}
\label{fig:cartpole_additional}
\end{figure}

We provide additional results on the Cartpole Swingup environment. In Figure~\ref{fig:cartpole_additional}(a), we fix the history buffer batch size $B=5$ and study the effect of $\pinit$ in the default version. We observe that $\pinit=0.0$ is a bad choice, as discussed in Section~\ref{sec:bsuite}. We also observe that for BootDDQN and DDQN-Bonus, the performance is comparable for $\pinit$ values in $[0.2, 1.0]$. This is due to the fact that in the reward is not very sparse in this version and thus local access does not lead to significant improvement. On the other hand, for PI-Bonus, local access leads to significant improvement over online access, and the best performance is achieved with a relatively small but non-zero $\pinit$, i.e., $0.2$. In Figure~\ref{fig:cartpole_additional}(b), we fix $\pinit=0.2$ and study the effect of the history buffer batch size $B$ for the default version of the environment. Again for BootDDQN and DDQN-Bonus, since local access does not lead to significant improvement, the performance is comparable for different values of $B$. But for PI-Bonus, choosing an uncertain element, i.e., $B>1$ is important. In Figure~\ref{fig:cartpole_additional}(c), we fix $\pinit=0.2$ and study the effect of the history buffer batch size $B$ for the hard version of the environment. Interestingly, we found that for BootDDQN, the normalized AUCs for $B=1, 5, 25, 125$ are comparable. Note that when $B=1$, we ignore the uncertainty and simply choose a random element from the history buffer. This indicates that for BootDDQN in the hard version of Cartpole Swingup, the gain of local access mainly comes from directly starting from an intermediate state, i.e., skipping the simulator queries from the initial state to the chosen one, and the uncertainty is less important for this setting.

\subsection{Atari}\label{sec:additiona_exp_atari}
\begin{figure}[ht]
\centering
\includegraphics[width=.4\linewidth]{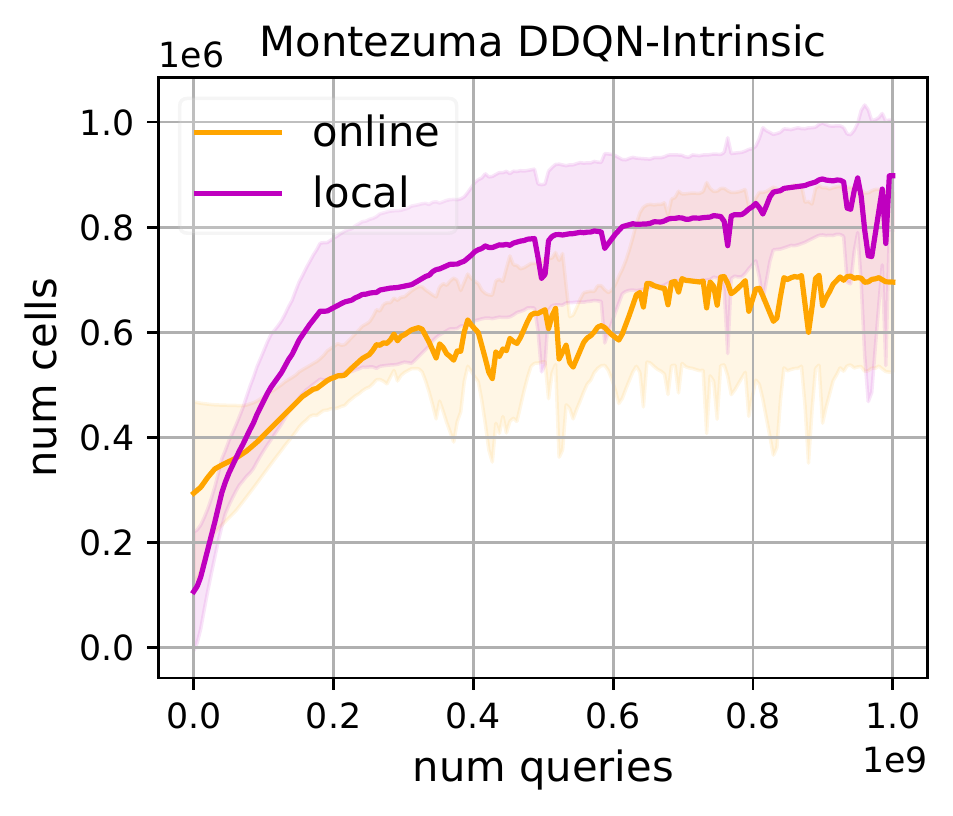}
\caption{The number of cells found by the DDQN-Intrinsic agent in the online and local settings.}
\label{fig:montezuma_num_cells_intrinsic}
\end{figure}

In Figure~\ref{fig:montezuma_num_cells_intrinsic}, we show the number of cells found by the DDQN-Intrinsic agent in the online and local settings. As we can see, with local access, the agent finds more cells than the online setting. This indicates that using local access, the agent can find a larger state space. On the other hand, comparing Figure~\ref{fig:montezuma_num_cells_intrinsic} and Figure~\ref{fig:montezuma_ablation}(a), we notice that using DDQN-Intrinsic and online access, we can find more cells than using DDQN and local access. However, DDQN with local access achieves better mean return. This implies that finding more cells created by the approximate-count based method does not necessarily lead to a better mean return.

\begin{table}[ht]
\centering
\caption{Evaluation results of DDQN-based agents on Montezuma's Revenge}
\scalebox{0.95}{
\begin{tabular}{c|c}
\toprule
Agent and access protocol & Result \\ \midrule
DDQN online & 2500.0 $\pm$ 0.0  \\ \hline
DDQN-Intrinsic online &  4980.0 $\pm$ 1086.91 \\  \hline
DDQN-Intrinsic local & 5518.0 $\pm$ 491.6  \\ \hline
DDQN local, approx count  & 6276.0 $\pm$ 551.3  \\ \hline
DDQN local, RND  & 5772.0 $\pm$ 955.75 \\
\bottomrule
\end{tabular}}
\label{table:eval_ddqn}
\end{table}

\begin{table}[ht]
\centering
\caption{Evaluation results, distributional DDQN}
\begin{tabular}{c|c|c|c}
\toprule
Game & Online & Local approx count & Local RND \\ 
\midrule
Montezuma's Revenge & 4421.33 $\pm$  869.63 & 6235.33 $\pm$ 860.91 & 5534.0 $\pm$ 118.11 \\
\hline
PrivateEye & 88258.27 $\pm$ 15908.99	& 100645.17 $\pm$ 62.16 &	100571.25 $\pm$ 93.18  \\
\hline
Venture & 1792.67 $\pm$ 37.39 & 1803.33 $\pm$ 50.34 & 1691.33 $\pm$ 33.4 \\
\hline
Pitfall & -1.31 $\pm$ 0.91 & 0.0 $\pm$ 0.0 & -2.21 $\pm$ 2.29 \\
\bottomrule
\end{tabular}
\label{table:eval_dist}
\end{table}

We report the evaluation results of the DDQN-based agents in Table \ref{table:eval_ddqn} and distributional DDQN in Table \ref{table:eval_dist}. For each seed, we consider the final performance of the agent to be the average of its last 30 evaluation runs. We then average this over 5 seeds and report this with 95\% confidence intervals. The number of environment steps is $10^9$ for Montezuma's revenge, and $2\times 10^8$ for the other Atari environments. We also report the maximum screen score that was seen during the acting over all seeds in Tables \ref{table:actor_ddqn} and \ref{table:actor_dist}. We calculate this by considering the screen score of a state as the sum of unclipped rewards attained till that state. Note that this score need not have been attained in a single data collection iteration in the local access setting. The agent could reset to a intermediate state with positive screen score, and then obtain the maximum screen achieved so far after several further steps.

On Montezuma's Revenge, the distributional DDQN local agent with approximate count uncertainty reaches a screen score $14300$. 
We note that we also observe screen scores $> 100000$ on this environment, but believe this may be due to triggering a treasure room curse bug, also mentioned in \citet{ecoffet2019go}. We also note that at least one of the RND local planning agent seeds reach a positive screen score on Pitfall.

\begin{table}[ht]
\centering
\caption{Maximum screen score during data collection, DDQN-based agents on Montezuma's Revenge}
\begin{tabular}{c|c}
\toprule
Agent and access protocol & Result \\
\midrule

DDQN online & 3000 \\ \hline
DDQN-Intrinsic online  & 6600   \\ \hline
DDQN-Intrinsic local & 8900  \\ \hline
DDQN local, approx count  & 9500 \\ \hline
DDQN local, RND  & 8000 \\

\bottomrule
\end{tabular}
\label{table:actor_ddqn}
\end{table}

\begin{table}[ht]
\centering
\caption{Maximum screen score during data collection, distributional DDQN}
\begin{tabular}{c|c|c|c}
\toprule
Game & Online & Local approx count & Local RND \\ 
\midrule
Montezuma's Revenge & 6000  & 14300 & 6500 \\
\hline
PrivateEye & 100800 & 100800 & 100800 \\
\hline
Venture & 2200 & 3000 & 2600 \\
\hline
Pitfall & 0 & 0  & 1800 \\
\bottomrule
\end{tabular}
\label{table:actor_dist}
\end{table}

\section{Checkpointing and Restoring the Environment}\label{sec:checkpointing}

For \bsuite~environments, since the simulators are implemented in Python, we can use the deep copy function to checkpoint and restore the environment state.

\begin{lstlisting}[language=Python,basicstyle=\small]
import copy

def make_env_checkpoint(env):
  return copy.deepcopy(env)

def restore_env(env_copy):
  return copy.deepcopy(env_copy)
\end{lstlisting}

For Atari games, we use the environment loader in the open-sourced Acme framework.\footnote{\href{https://github.com/deepmind/acme/blob/master/examples/baselines/rl\_discrete/helpers.py\#L37}{https://github.com/deepmind/acme/blob/master/examples/baselines/rl\_discrete/helpers.py\#L37}}
We use the ``clone'' and ``restore'' function of the environment to make checkpoints and restore the states. Note that since we stack the last 4 frames of the game, we also need to checkpoint and restore the frame stack.

\begin{lstlisting}[language=Python,basicstyle=\small]
import collections
import copy
import numpy as np

def make_env_checkpoint(env):
  state=env.unwrapped.clone_full_state()
  stacker_state=copy.deepcopy(np.array(env._frame_stacker._stack))
  stacker_max_size=env._frame_stacker._stack.maxlen
  return state, stacker_state, stacker_max_size

def restore_env(env, state, stacker_state, stacker_max_size):
  env.unwrapped.restore_full_state(element.state)
  env._environment._frame_stacker._stack = collections.deque(
    copy.deepcopy(element.stacker_state), 
    maxlen=element.stacker_max_size)
  return env
\end{lstlisting}

\section{Hyperparameters}\label{sec:hprams}

\subsection{Deep Sea}\label{sec:hparams_deep_sea}

For all the Q-networks used in our experiments on Deep Sea, we use an MLP with two hidden layers, each with size $64$.

For the BootDDQN agent, we use an ensemble of size $20$ and a prior scale~\citep{osband2018randomized} of $40.0$. For all other agents, we use covariance-based uncertainty metric $u_{\text{cov}}$ with random Fourier features. We choose feature dimension $d=1500$. For DDQN and DDQN-Bonus, we choose regularization coefficient $\lambda = 0.01$ and for PI-Bonus, we choose $\lambda = 0.1$. For DDQN-Bonus and PI-Bonus, we use bonus scale $c=1.0$. In all the experiments, we use a replay buffer of size $10^6$, sufficient to store all the transitions in the experiments. All other hyperparameter are provided in Table~\ref{table:deep_sea_hparams}. In the table, SGD period means the number of new observations we obtain before starting a new SGD training step; environment checkpoint period means the number of observations we obtain before saving a new environment checkpoint in the history buffer.

\begin{table}[ht]
\centering
\caption{Hyperparameters for Deep Sea experiments}
\begin{tabular}{c | c }
\toprule
Hyperparameter & Value \\
\midrule
discount $\gamma$ & $1.0$ for PI-Bonus, $0.99$ for other agents \\ \hline
$\epsilon$-greedy & $0.1$ for DDQN-Bonus, $0.0$ for other agents \\ \hline
SGD period & $4$ for PI-Bonus, $1$ for all other agents  \\  \hline
target update period  & $4$ (N/A for PI-Bonus) \\  \hline
environment checkpoint period & $1$ \\ \hline
learning batch size & $128$ \\  \hline
optimizer & Adam \\ \hline
learning rate & $0.001$ \\  \hline
maximum gradient norm for clipping & $20$ \\
\bottomrule
\end{tabular}
\label{table:deep_sea_hparams}
\end{table}

\subsection{Cartpole Swingup}\label{sec:hparams_cartpole}

\paragraph{Environment parameters} Recall that in Figure~\ref{fig:bsuite}, the horizontal position of the cart is denoted by $x$, the angle between the pole and the upright direction is denoted by $\alpha$. We also denote the angular velocity of the pole by $\dot \alpha$. In the default version, the agent receives a positive reward if the following conditions are satisfied:
\[
\begin{cases}
&|x| < 1.0 \\
& \cos\alpha >0.5 \\
&|\dot \alpha| < 1.0,
\end{cases}
\]
and in the hard version, the agent receives a positive reward if the following conditions are satisfied:
\[
\begin{cases}
&|x| < 0.05 \\
& \cos\alpha >0.95 \\
&|\dot \alpha| < 1.0.
\end{cases}
\]
Therefore, the reward is more sparse in the hard version.

\paragraph{Agent hyperparameters}

For all the Q-networks used in our experiments on Cartpole Swingup, we use an MLP with two hidden layers, each with size $128$.

For BootDDQN, we sweep different combinations of ensemble size and prior scale to determine the best combination and then use it in the local setting. For the default version, we use an ensemble of size $5$ and prior scale of $5.0$. For the hard version, we use an ensemble of size $10$ and prior scale of $5.0$. We use $\epsilon=0.2$ for BootDDQN.

For DDQN-Bonus and PI-Bonus, we use random Fourier feature of dimension $d=500$ and regularization coefficient $\lambda=0.1$. For DDQN-Bonus, we sweep the bonus scale parameter $c$ and the $\epsilon$-greedy parameter in the online setting and use the best combination in the local setting. For DDQN-Bonus, in the default version of Cartpole Swingup, we use bonus scale $c=10$ and $\epsilon=0.2$, and in the hard version of Cartpole Swingup, we use bonus scale $c=100$ and $\epsilon=0.0$. For PI-Bonus, we use bonus scale $c=500$ and $\epsilon=0.0$.

As for the hyperparameters for local planning, we use a history buffer of size $10^5$. We sweep different combinations of $\pinit$ and history buffer batch size $B$ and choose the best combination for each agent. The hyperparameters that we choose for results in Figure~\ref{fig:cartpole_swingup}(a, b, c) are given in Table~\ref{table:cartpole_local_hparams}.

\begin{table}[ht]
\centering
\caption{Local planning hyperparameters in Figure~\ref{fig:cartpole_swingup}.}
\begin{tabular}{c | c | c | c }
\toprule
Agent & Version & $\pinit$ & $B$ \\
\midrule
BootDDQN & default & $0.8$ &  $5$  \\ \hline
DDQN-Bonus & default & $0.2$ & $5$  \\ \hline
PI-Bonus & default & $0.2$ & $5$ \\  \hline
BootDDQN & hard & $0.4$ &  $125$  \\ \hline
DDQN-Bonus & hard & $0.2$  & $25$ \\
\bottomrule
\end{tabular}
\label{table:cartpole_local_hparams}
\end{table}

All other hyperparameters are provided in Table~\ref{table:cartpole_hparams}. 

\begin{table}[ht]
\centering
\caption{Hyperparameters for Cartpole Swingup experiments}
\resizebox{\columnwidth}{!}{
\begin{tabular}{c | c }
\toprule
Hyperparameter & Value \\
\midrule
discount $\gamma$ & $0.99$ for BootDDQN and DDQN-Bonus, $0.995$ for PI-Bonus \\ \hline
SGD period & $25$ for BootDDQN and DDQN-Bonus, $5$ for PI-Bonus  \\  \hline
target update period  & $10$ (N/A for PI-Bonus) \\  \hline
environment checkpoint period & $5$ \\ \hline
learning batch size & $2048$ for BootDDQN and DDQN-Bonus, $256$ for PI-Bonus \\  \hline
optimizer & Adam \\ \hline
learning rate & $0.001$ \\  \hline
maximum gradient norm for clipping & $20$ \\
\bottomrule
\end{tabular}
}
\label{table:cartpole_hparams}
\end{table}

\subsection{Atari}\label{sec:hparams_atari}

\paragraph{Environment parameters}

We run experiments on the v0 version of Atari environments with sticky actions, making the environments stochastic \citep{machado2018revisiting}.
We apply the standard frame processing wrapper provided in Acme,\footnote{\href{https://github.com/deepmind/acme/blob/master/acme/wrappers/atari\_wrapper.py}{https://github.com/deepmind/acme/blob/master/acme/wrappers/atari\_wrapper.py}} which includes converting the frames to grayscale, downsampling, stacking $4$ consecutive frames for each observation. We summarize the environment hyperparameters in Table~\ref{table:atari_env}. 

\begin{table}[ht]
\centering
\caption{Hyperparameters for Atari environments.}
\begin{tabular}{c | c }
\toprule
Hyperparameter & Value \\
\midrule
max episode length & $30$ min ($108,000$ steps) \\ \hline
number of stacked frames & $4$  \\ \hline
zero discount on life loss & false  \\ \hline
random noops range $30$ & not used \\ \hline
sticky actions & true, repeat action probability $0.25$ \\ \hline
frame size & $(84, 84)$ \\ \hline
grayscaled/RGB & grayscaled  \\ \hline
action set & full \\
\bottomrule
\end{tabular}
\label{table:atari_env}
\end{table}

\paragraph{Agent hyperparameters}

As for model architecture, for both DDQN-based agents and distributional DDQN, we use the Acme AtariTorso network architecture\footnote{\href{https://github.com/deepmind/acme/blob/master/acme/jax/networks/atari.py}{https://github.com/deepmind/acme/blob/master/acme/jax/networks/atari.py}} followed by an MLP with 512 hidden units.

As for local planning hyperparameters, we use a history buffer of size $10^6$. Under local access, we reset to the highest-uncertainty state in the sampled batch, and then take a random action, i.e., Eq.~\eqref{eq:most_uncertain_2}. For the DDQN-based agents in Figure~\ref{fig:montezuma_ddqn}, for local access runs, we sweep different combinations of $\pinit$ and $B$ and report the results corresponding to the best combination we can find. The values of $\pinit$ and $B$ that we use in the experiments in Figure~\ref{fig:montezuma_ddqn} are provided in Table~\ref{table:montezuma_ddqn_hparam}.
For the DDQN agent in Montezuma's Revenge, we find that using $\epsilon=0.1$ leads to better results in the online setting, and thus we choose $\epsilon=0.1$ for this setting. We use $\epsilon=0.01$ for all other settings.
For the distributional DDQN agent, the local access runs in Figure~\ref{fig:distddqn}, we use $\pinit=0.3$ and history buffer batch size $B=32$.

\begin{table}[ht]
\centering
\caption{Local planning hyperparameters in Figure~\ref{fig:montezuma_ddqn}.}
\begin{tabular}{c | c | c }
\toprule
Agent  & $\pinit$ & $B$ \\
\midrule
DDQN local approximate count  &  $0.2$  &  $8$  \\ \hline
DDQN-Intrinsic local approximate count  & $0.7$  & $8$  \\ \hline
DDQN local RND   &  $0.3$  &  $32$ \\ 
\bottomrule
\end{tabular}
\label{table:montezuma_ddqn_hparam}
\end{table}

In Montezuma's Revenge, in order to limit online access queries which do not lead to significant exploration, and also have more exploration around uncertain regions, we limit the maximum length of all roll-outs to $2000$. Another possible benefit of this is the composition of the limited size replay buffer is more uniform, and not dominated by certain long rollouts from unimportant regions.
For all other settings, including both online and local access for other games, and the online access setting for Montezuma's Revenge, we did not find this option helpful. Thus, we run rollout until the end of the episode for all other settings.

The remaining hyperparameters corresponding to the Acme DDQN config\footnote{\href{https://github.com/deepmind/acme/blob/master/acme/agents/jax/dqn/config.py}{https://github.com/deepmind/acme/blob/master/acme/agents/jax/dqn/config.py}} are given in Table~\ref{table:distddqn}. Compared to the default settings, we do not use prioritized replay sampling, we clip gradients, and we use higher ``samples per insert'', which governs the average number of times a transition is sampled by the learner before being evicted from the replay buffer. We did not find prioritized sampling to help the performance of either the online or local access agent on Montezuma's Revenge. For the training system, we use $64$ CPU machines as actors and $1$ TPU machine as the learner in most settings. The only exception is that for RND we use $128$ actors in order to speedup training. We did not find using $128$ actors improving the online baseline.

\begin{table}[ht]
\centering

\caption{Hyperparameter values for experiments on Atari games.}
\resizebox{\columnwidth}{!}{
\begin{tabular}{c | c } 
\toprule
Hyperparameter & Value \\
\midrule
discount $\gamma$ & $0.997$ for PrivateEye, $0.99$ for all other games \\ \hline
learning batch size & $256$ for distributional DDQN, $128$ for DDQN and DDQN-Intrinsic \\  \hline
\multirow{2}{*}{learning rate}  & $10^{-4}$ for DDQN and DDQN-Intrinsic \\
       & chosen between  $10^{-4}$ and $2 \times 10^{-5}$ for distributional DDQN
\\ \hline
epsilon & $0.1$ for DDQN online in Montezuma's Revenge, $0.01$ otherwise \\  \hline
eval epsilon & $0$ \\  \hline
Adam epsilon & $10^{-5}$\\  \hline
number of TD steps (n\_step) & $5$ for distributional DDQN, $3$ for DDQN and DDQN-Intrinsic \\  \hline
target update period & $2500$ \\  \hline
environment checkpointing period & $10$  \\ \hline
number of actors & $128$ for experiments with RND, $64$ otherwise \\ \hline
max gradient norm & $10$ \\  \hline
min replay size & $5\times 10^4$  \\  \hline
max replay size & $5\times 10^6$  \\  \hline
importance sampling exponent & $0$ \\  \hline
priority exponent & $0$  \\  \hline
samples per insert & $8$  \\  \hline
number of SGD steps per step & $8$ \\  \hline
number of atoms in distributional DDQN & $201$  \\  \hline
$v_{\min}$ in distributional DDQN & $-1$  \\ \hline
$v_{\max}$ in distributional DDQN & $120000$ for PrivateEye, $20000$ for all other games \\ \hline
$v_{\max}$ in distributional DDQN RND & $120000$ for PrivateEye, $10000$ for all other games \\ \hline
bonus scale $c$ in DDQN-Intrinsic  & $0.1$ \\  \hline
reward clipping &  $1$ for DDQN and DDQN-Intrinsic, N/A for distribution DDQN \\
\bottomrule
\end{tabular}
}
\label{table:distddqn}
\end{table}

\end{document}